\documentclass[10pt,twocolumn,letterpaper]{article}

\usepackage{wacv}
\usepackage{times}
\usepackage{epsfig}
\usepackage{graphicx}
\usepackage{amsmath}
\usepackage{amssymb}
\usepackage{booktabs}
\usepackage{xcolor}
\usepackage[normalem]{ulem}
\usepackage{here}
\usepackage{multirow}
\usepackage[accsupp]{axessibility}

\def\wacvPaperID{332} % Enter the WACV Paper ID here

\wacvalgorithmstrack   % Uncomment this line if you are submitting to the Algorithms Track.
\wacvfinalcopy % *** Uncomment this line for the final submission

\ifwacvfinal
\usepackage[breaklinks=true,bookmarks=false]{hyperref}
\else
\usepackage[pagebackref=true,breaklinks=true,colorlinks,bookmarks=false]{hyperref}
\fi

\begin{document}

%%%%%%%%% TITLE
\title{Ego-Vehicle Action Recognition based on Semi-Supervised Contrastive Learning}

% \author{Chihiro Noguchi\\
% InfoTech, Connected Advanced Development Div.\\
% Connected Company, TOYOTA Motor Corporation, Japan\\
% {\tt\small chihiro_noguchi_aa@mail.toyota.co.jp}
% % For a paper whose authors are all at the same institution,
% % omit the following lines up until the closing ``}''.
% % Additional authors and addresses can be added with ``\and'',
% % just like the second author.
% % To save space, use either the email address or home page, not both
% \and
% Toshihiro Tanizawa\\
% InfoTech, Connected Advanced Development Div.\\
% Connected Company, TOYOTA Motor Corporation, Japan\\
% {\tt\small toshihiro_tanizawa@mail.toyota.co.jp}
% }
\author{
Chihiro Noguchi \qquad\qquad Toshihiro Tanizawa \\
Toyota Motor Corporation, Japan \\
{\tt\small \{chihiro\_noguchi\_aa, toshihiro\_tanizawa\}@mail.toyota.co.jp}
}

\maketitle
\thispagestyle{empty}

%%%%%%%%% ABSTRACT
\begin{abstract}
In recent years, many automobiles have been equipped with cameras, which have accumulated an enormous amount of video footage of driving scenes. Autonomous driving demands the highest level of safety, for which even unimaginably rare driving scenes have to be collected in training data to improve the recognition accuracy for specific scenes. However, it is prohibitively costly to find very few specific scenes from an enormous amount of videos. In this article, we show that proper video-to-video distances can be defined by focusing on ego-vehicle actions. It is well known that existing methods based on supervised learning cannot handle videos that do not fall into predefined classes, though they work well in defining video-to-video distances in the embedding space between labeled videos. To tackle this problem, we propose a method based on semi-supervised contrastive learning. We consider two related but distinct contrastive learning: standard graph contrastive learning and our proposed SOIA-based contrastive learning.
We observe that the latter approach can provide more sensible video-to-video distances between unlabeled videos.
Next, the effectiveness of our method is quantified by evaluating the classification performance of the ego-vehicle action recognition using HDD dataset, which shows that our method including unlabeled data in training significantly outperforms the existing methods using only labeled data in training.
\end{abstract}

%%%%%%%%% BODY TEXT

\begin{figure}[htbp!]
    \hspace{-0.55cm}
    \begin{tabular}{c}
    \begin{minipage}[t]{0.49\linewidth}
        \centering
      \includegraphics[clip, width=8.6cm]{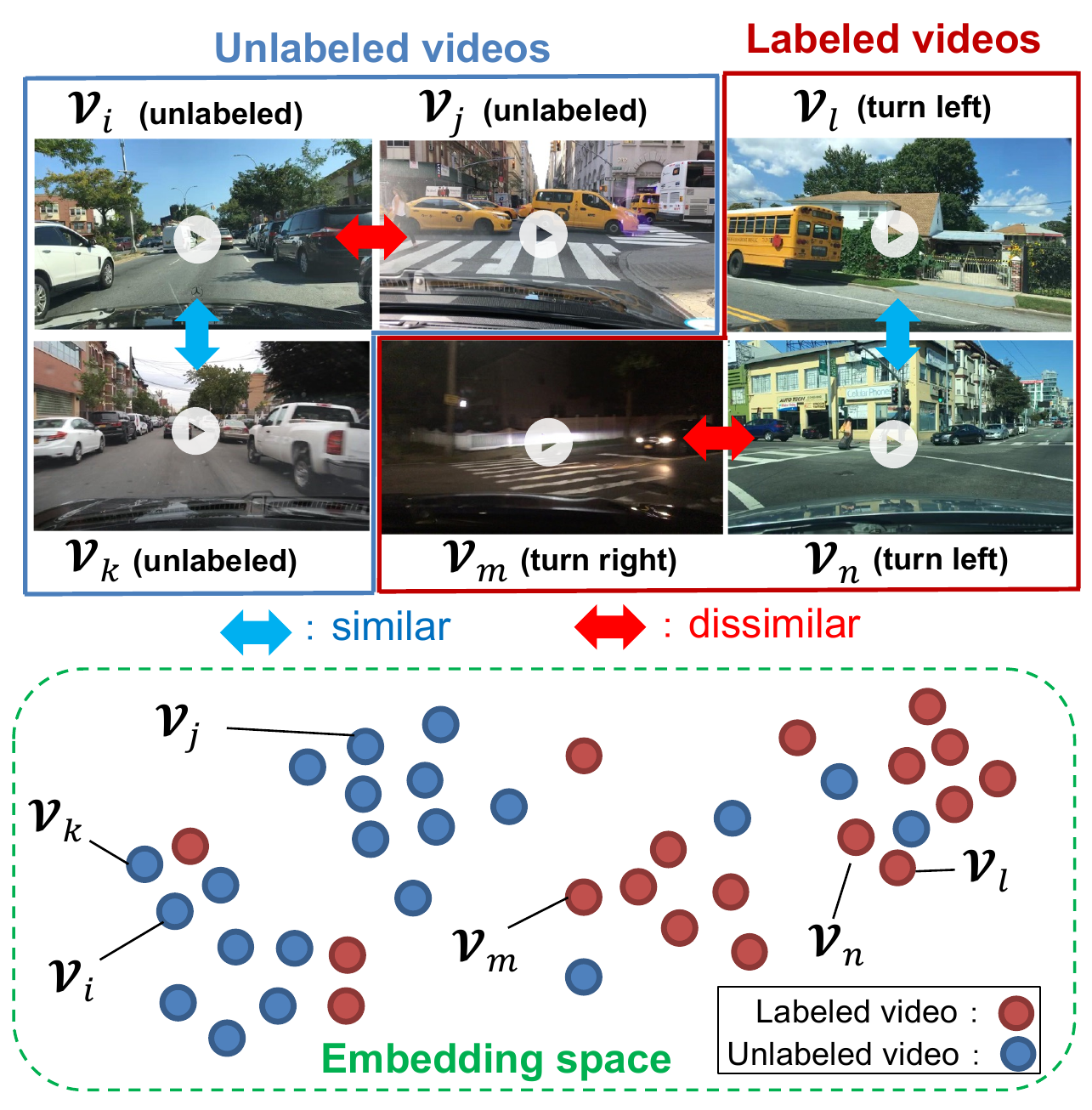}
      \end{minipage}
    \end{tabular}
    \caption{The goal of the proposed method is to set sensible distances in the embedding space between input driving videos. Labeled videos with the same label are embedded closer together while those with different labels are embedded farther apart. On the other hand, how to optimize the distances between unlabeled videos is not trivial. To address this problem, we use semi-supervised contrastive learning techniques in this article. As illustrated above, the proposed method can set sensible distances between all of both labeled and unlabeled videos.}
    \label{fig:embedding}
\end{figure}

\section{Introduction}
Autonomous driving technology has received an enormous amount of attention in recent years.
For autonomous driving to be realized, various challenging problems in computer vision have to be solved
\cite{BADUE2021113816}.
Many of the recent studies are based on deep learning models, which require a huge number of human annotated data for training. Since autonomous driving requires the highest level of safety, it is necessary to collect a sufficient number of labeled data even for situations in very rare occasions.
To collect video data for such rare scenes out of an enormous number of ordinary scenes
can be another challenging problem to be solved.
In this article, we focus on ego-vehicle action recognition which is a task to predict a scene-level label with respect to an action of the ego-vehicle from a video taken by the front camera of the ego-vehicle as a single input.
To extract relevant scenes that occur in rare occasions,
human annotators are required to go through a large number of irrelevant scenes, which can be a prohibitive cost.
If we are able to define a sufficiently meaningful distance between all obtained scenes including unlabeled ones, it would enable us to find relevant rare scenes systematically and even automatically (Fig. \ref{fig:embedding}).

To set proper distances between unlabeled videos, we adapt a methodology of contrastive learning (CL) \cite{9157636,pmlr-v119-chen20j}.
The CL is widely used for representation learning in various areas and enables us to obtain meaningful representations from unlabeled data.
In the current setting, we can readily obtain a large number of videos with scene-level labels that can be occurred frequently in ordinary driving scenes, such as left turns, U-turns, lane changes, and so on.
Therefore, it is more desirable to extend the standard CL defined in an unsupervised setting to in a semi-supervised setting in order to set distances between all of both types of videos.
Recently, Khosla et al.\ \cite{NEURIPS2020_d89a66c7} proposed supervised CL, which extends the standard CL to in a fully-supervised setting.
In this article, we further extend this framework of CL to the current semi-supervised context.
Hence the semi-supervised CL (SSCL).

A simple approach to extract relevant features from video data is to feed them into a convolutional neural network (CNN) to learn scene-level labels.
This approach is called the object-agnostic method \cite{Wang2019DeepOP}.
However, despite that this approach takes advantage of all the information contained in the original videos,
it does not use the ground-truth annotations for object instances.
As a result, according to \cite{Wang2019DeepOP}, the object-centric method, which uses both object instances and end-to-end learning,
outperforms the object-agnostic method in the same settings.
Based on this understanding, we take the object-centric approach in the method proposed in this article.
More specifically, input video data are transformed into graph structures by constructing spatio-temporal graphs (ST-graphs) \cite{Wang_2018_ECCV}. Nodes in the ST-graph represent objects such as cars and pedestrians, and edges represent spatial and temporal relationships between the objects.
Then the ST-graphs are fed into a graph convolutional network (GCN) \cite{Kipf:2016tc,Patil_2021_CVPR,Wang2021TrackWA} to obtain feature vectors of videos.

CL can also be applied to data with graph structures, which is called graph CL (GCL) \cite{you2020graph,10.1145/3404835.3462862}.
In the GCL, a positive sample is generated from data augmentation such as node dropping, edge perturbation and attribute masking \cite{you2020graph}.
From the framework of GCL, it might be expected that unlabeled ST-graphs with similar structures are mapped close to each other in the embedded space.
In practice, however, we observed that such na\"{i}ve distances between unlabeled videos in the embedded space often deviate from the ones that are understandable in an ordinary sense.
To address this problem, we propose a Simple algorithm for Object Instance Association (SOIA), which is inspired by Simple Online and Realtime Tracking (SORT), proposed in \cite{7533003}. It provides distances between videos by associating the detected object instances between different driving videos and calculating IoU-based distances.
The distances determined by SOIA (SOIA distances) can be applied to the CL, where a positive sample is determined as the closest sample to an anchor sample in a batch, while negative samples are the remainder.
We observed that the SSCL with the SOIA distances can provide more sensible distances.
We call this approach SOIA-based CL.

It is difficult to quantitatively evaluate the quality of the obtained distances between all of videos including labeled and unlabeled videos.
On the other hand, only labeled videos can be evaluated quantitatively.
Therefore, we consider two kinds of evaluation methods.
For unlabeled videos, the quality of the distances is evaluated by showing many examples of video pairs of the query and its closest videos in the embedding space.
For labeled videos, we evaluate the classification performance in terms of the ego-vehicle action recognition.
In the evaluation on the labeled videos, the proposed method in this article achieves the state-of-the-art performance compared to existing methods trained in a fully-supervised setting.
Our experiments show that the performance of the proposed method is significantly improved by using the unlabeled videos as well for training.

Our contribution is summarised as follows.
(1) We propose a framework to provide sensible distances between driving videos even when they are unlabeled. To this end, we propose SOIA to determine distances between unlabeled videos and use the distances following the SSCL manner. (2) Our method proposed in this article achieves the state-of-the-art performance in the ego-vehicle action recognition. The performance is significantly improved by using unlabeled as well as labeled videos combined with SSCL techniques.

\section{Related Works}

\subsection{Ego-Vehicle Action Recognition}
A number of methods have been proposed for recognizing ego-vehicle actions, including those that use sensor signals or video images, and both of them.
The Hidden Markov model (HMM) is a common approach that uses sensor signals \cite{Oliver2000GraphicalMF,Kuge2000ADB,1438388,7410721}.
A hidden state corresponds to a scene-level label, and they predict the transition of the hidden states using the input signals.
Another common approach is to use recurrent neural networks \cite{8099859,8578901,9009797}.
They extract image features by using convolutional layers, and temporal relationships with neighboring frames are embedded in the features by using long short-time memory.
Promising methods that are recently proposed
include spatio-temporal (ST) graph construction from input videos.
ST-graph construction is used for a variety of recognition tasks, such as human actions \cite{yan2018spatial} and group activities \cite{Wu_2019_CVPR}, not limited to ego-vehicle actions.
After constructing ST-graphs, their spectral features are used to train classifiers \cite{Chandra2020CMetricAD,Chandra2020GraphRQICD}.
Graph neural networks are also widely used in extracting the deeper features
\cite{9022086,Li2020Learning3E,Mylavarapu2020UnderstandingDS}.

\subsection{Semi-Supervised Contrastive Learning (SSCL)}
The objective of semi-supervised learning (SSL) is to utilize readily available unlabeled data for helping classifiers improve performance and reducing reliance on labeled data.
It has achieved remarkable results in computer vision by introducing various techniques, including entropy minimization \cite{NIPS2004_96f2b50b,lee2013pseudo}, Mean Teacher \cite{Tarvainen2017MeanTA,Xie_2020_CVPR}, MixMatch \cite{NEURIPS2019_1cd138d0,berthelot2019remixmatch,NEURIPS2020_06964dce},  consistency regularization \cite{Bachman2014LearningWP,10.5555/2969442.2969635,DBLP:conf/iclr/LaineA17,NEURIPS2020_44feb009}, and label propagation \cite{pmlr-v80-kamnitsas18a}.
However, these methods basically assume that each of unlabeled data has an unobserved label that belongs to predefined classes.
Therefore, their performance is significantly degraded when the unlabeled data contains Out-of-Distribution (OoD) data \cite{10.5555/3327144.3327244}.
To tackle this problem, recent studies introduce OoD filters that classify OoD and in-distribution data, using predictive uncertainty \cite{Chen_Zhu_Li_Gong_2020}, empirical risk minimization \cite{pmlr-v119-guo20i}, and distance from class-wise prototypes \cite{park2021opencos}.
The proposed method in this article takes the setting in which unlabeled data contains only OoD data.

Contrastive learning (CL) \cite{9157636,pmlr-v119-chen20j} has attracted much attention due to its outstanding abilities for representation learning in computer vision.
The success of CL stems from an instance discrimination pretext task \cite{Wu2018UnsupervisedFL}, where each instance is attracted to its augmentation and other instances are drawn away from it in the embedding space.
Recently, Khosla et al. \cite{NEURIPS2020_d89a66c7} proposed supervised CL, which is a framework to apply CL to a fully-supervised setting.
The supervised CL considers two types of positive samples for an anchor sample: an augmented view of the anchor sample and samples with the same label as that of the anchor sample in a batch.
Cui et al. \cite{Cui_2021_ICCV} solved its difficulty of imbalance learning by introducing class-wise learnable prototypes.
Recently, CL has also been applied to semi-supervised setting \cite{Li2021CoMatchSL,park2021opencos,9732218}.
They generate pseudo-labels for unlabeled data so that they can be handled by the usual CL.
Contrasting to this line of works, the proposed method here is applied to the semi-supervised setting by extending the ordinary supervised CL \cite{Cui_2021_ICCV}.

\begin{figure*}[htbp!]
    \hspace{-0.5cm}
    \begin{tabular}{c}
    \begin{minipage}[t]{0.95\linewidth}
        \centering
      \includegraphics[clip, width=17.5cm]{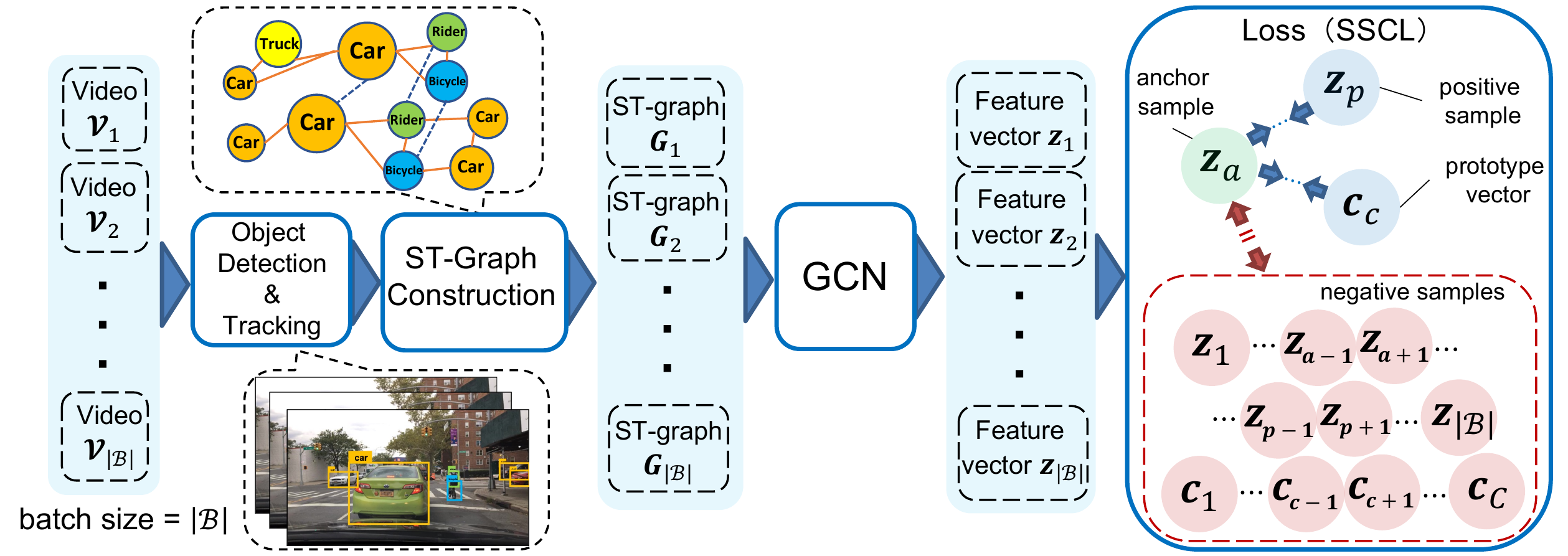}
      \end{minipage}
    \end{tabular}
    \caption{Overview of the proposed method. First, object instances in input videos are detected by any object detection and tracking methods. Then ST-graphs are constructed from the detected object instances. These ST-graphs are input to a GCN, and feature vectors for each ST-graph are obtained. Finally, the GCN is optimized using the loss function of SSCL. This figure illustrates SOIA-based CL. The sample closest to an anchor sample in a batch in terms of the SOIA distance is regarded as a positive sample, while samples, which are neither anchor nor positive, are negative. In addition, only for labeled samples, anchor sample is optimized to be close to the corresponding learnable prototype vector and be far away from the other prototype vectors (see Sec. \ref{sec:gcn_training} for more details).}
    \label{fig:overview}
\end{figure*}

\section{Method}
The overview of our approach is illustrated in Fig. \ref{fig:overview}.
First, input driving videos are transformed into ST-graphs by using bounding boxes and semantic labels of detected objects, which is described in Sec.\ \ref{sec:st_graph}.
Second, the ST-graphs are fed into a GCN. This architecture is described in Appendix A.
Third, the GCN is trained through SSCL. This process is described in Sec.\ \ref{sec:gcn_training}.

\subsection{Spatio-Temporal (ST) graph construction}
\label{sec:st_graph}
In this subsection, we describe how to construct an ST-graph $G_n=(V_n, E_n)$ from a video $n$,
where $V_n$ denotes a set of nodes corresponding to objects detected in each frame of the video $n$,
and $E_n$ denotes a set of edges corresponding to spatial or temporal weights between the objects.
The objects in $V_n$ can be acquired from the video $n$ by using any object tracking methods.
Note that we use two terms, ``object'' and ``object instance'', differently. An ``object'' stands for an individual object in each image, while an ``object instance'' corresponds to
a set of objects with the same ID in a series of frames of a video.
We denote $\mathcal{I}_n$ as a set of object instances in the video $n$.

\noindent
\textbf{Node Attributes.}\quad
To feed the initial graph to a GCN system, we have to assign a suitable set of node attributes.
We set three types of node attributes: (1) semantic labels,
(2) geometric features of bounding boxes, and (3) interaction with lane lines.
A semantic label $\boldsymbol{s}_i\in\mathbb{R}^8$ is a one-hot vector for each bounding box $i$.
According to \cite{Yu2020BDD100KAD}, we include eight object classes: \textit{pedestrian}, \textit{rider}, \textit{car}, \textit{truck}, \textit{bus}, \textit{train}, \textit{motorcycle} and \textit{bicycle}.
A geometric feature $\boldsymbol{g}_i$ is defined as
$\boldsymbol{g}_i=\left(
\frac{a_i}{W}, \frac{b_i}{H}, \frac{w_i}{W}, \frac{h_i}{H}, \frac{w_ih_i}{\sqrt{WH}}
\right), \forall i\in V_n$,
where $(a_i,b_i)$, $w_i$ and $h_i$ denotes the centroid coordinate, the width and height of bounding box $i$, respectively. $W$ and $H$ denote the overall width and height of the input video clip.
Finally, information of lane lines on the road is also essential to recognize ego-vehicle actions. To incorporate it into the model, we introduce the interaction between objects and lane lines. More specifically, we consider the interaction with the five points of bounding box $i$, $a_1^i, a_2^i, a_3^i, a_4^i, a_5^i$ (the four corners and the center), and all pixels corresponding to lane lines in an image. 
In the following, superscript $i$ is omitted for readability.
The interaction between $a_k$, $\forall k\in\{1,\dots,5\}$ and lane line pixels is defined as
$\boldsymbol{f}_{a_k}=\sum_{p\in\mathcal{M}}\frac{w_p^{a_k}\boldsymbol{v}_p^{a_k}}{\|\boldsymbol{v}_p^{a_k}\|_2}, \forall k\in\{1,\dots,5\}$,
where $w_p^{a_k}=e^{-\frac{\left(d_p^{a_k}\right)^2}{2\sigma^2}}$. $\mathcal{M}$ denotes a set of pixels corresponding to lane lines in an image. $\boldsymbol{v}_p^{a_k}\in\mathbb{R}^2$ is 2D-vector from $a_k$ to $p$, and $d_p^{a_k}$ indicates the Euclidean distance between the two points $a_k$ and $p$. 
Introducing weight $w_p^{a_k}$ induces that pixels closer to the bounding box are more strongly affected.
As a result, for bounding box $i$, lane line features $\boldsymbol{f}_i\in\mathbb{R}^{10}$ are obtained by concatenating as $\boldsymbol{f}_i=[\boldsymbol{f}_{a_1}, \boldsymbol{f}_{a_2}, \boldsymbol{f}_{a_3}, \boldsymbol{f}_{a_4}, \boldsymbol{f}_{a_5}]$.

\noindent
\textbf{Edge Weights.}\quad
$E_n$ contains two types of edges: spatial and temporal edges.
The spatial edges represent spatial similarity between bounding boxes within a frame, and the temporal edges represent temporal relationships between adjacent frames.
Edge weight $e_{ij}$, $\forall(i,j)\in E_n$ is defined as follows:
\begin{align}
    e_{ij} = \left\{
\begin{array}{ll}
e^{-\frac{d^2_{ij}}{2\sigma^2}} & (\mathrm{if}\ e_{ij}\ \mathrm{is}\ \mathrm{a}\ \mathrm{spatial}\ \mathrm{edge})\\
1 & (\mathrm{if}\ e_{ij}\ \mathrm{is}\ \mathrm{a}\ \mathrm{temporal}\ \mathrm{edge}) \\
0 & (\mathrm{otherwise})
\end{array}
\right.
\end{align}
where $d_{ij}$ denotes the Euclidean distance between the centroids of bounding box $i,j$, and $\sigma=\sqrt{H^2+W^2}/4$.
$e_{ij}$ denotes a spatial edge if two bounding boxes $i$ and $j$ are in the same frame.
A temporal edge exists only if $i$ and $j$ correspond to the same object instance between two adjacent frames.
As a result, the nodes are densely connected within a frame, while the entire ST-graph is a sparse graph.

\subsection{GCN Training}
\label{sec:gcn_training}
GCN training in the semi-supervised setting uses both labeled ST-graphs $\mathcal{D}_l=\{G_n, y_n\}_{n=1}^{N_l}$ and unlabeled ST-graphs $\mathcal{D}_u=\{G_n\}_{n=N_l+1}^{N_l+N_u}$, where $y_n$ denotes a scene-level label of video $n$, and $N_l$ and $N_u$ denote the number of labeled and unlabeled ST-graphs, respectively ($N=N_l+N_u$). ST-graph $n$ is obtained from video $n$, as explained in Sec. \ref{sec:st_graph}.

For the GCN training, we consider two types of approaches: standard graph contrastive learning (GCL) and SOIA-based CL (SCL). The difference between the two approaches lies in how the positive and negative samples are generated. The following Sec. \ref{sec:graph_contrastive_learning} and \ref{sec:soia} describe how to generate the positive and negative samples in the two approaches, respectively. And then, Sec. \ref{sec:semi_supervised_contrastive_learning} describes the detail of SSCL and defines the loss function.

\subsubsection{Graph Contrastive Learning (GCL)}
\label{sec:graph_contrastive_learning}
GCL deals with data with graph structures. To generate a positive sample, data augmentation considering the graph structures is applied to the anchor sample. Following \cite{you2020graph}, we adopt three data augmentations: node dropping, edge perturbation and attribute masking.
As in a normal CL, samples other than the anchor sample in a batch are used as negative samples.

\subsubsection{SOIA-based Contrastive Learning (SCL)}
\label{sec:soia}
The goal of the SOIA is to associate object instances across difference videos. After the association, IoU-based distances can be measured between the associated object instances.
As a result, distances between videos can be defined based on the IoU-based distances.
In SOIA-based CL, a positive sample is found as the closest sample in a batch to the anchor sample in terms of the SOIA distance, and negative samples are samples other than the positive and anchor samples in a batch. In the following, the detail of the SOIA and how to select positive and negative samples are explained.

\noindent
\textbf{Simple algorithm for Object Instance Association (SOIA).}\quad
The SOIA is inspired from SORT \cite{7533003}, which is a popular algorithm for data association in object tracking. The SORT associates objects across different two frames to detect an object instance, while the SOIA associates object instances across different two videos to define a distance between the two videos.

We first measure similarities between object instances across different ST-graphs.
$S_{nm}\in\mathbb{R}^{|\mathcal{I}_n|\times |\mathcal{I}_m|}$ is a similarity matrix that indicates object instance similarities across ST-graphs $n$ and $m$.
In other words, its $(u,v)$th element $s_{nmuv}$, $\forall u\in\mathcal{I}_n$, $\forall v\in\mathcal{I}_m$, indicates the similarity between object instances $u$ and $v$.
The similarity $s_{nmuv}$ is determined by mean intersection over union (mIoU) between bounding boxes of objects in $\mathcal{N}_{nu}$ and $\mathcal{N}_{mv}$. Here, $\mathcal{N}_{nu}$ denotes a set of objects corresponding to an object instance $u$ in ST-graph $n$.
Hence,
$s_{nmuv}=\frac{1}{T}\sum_{t=1}^T \mathrm{IoU}(u_t, v_t),$
where $T$ denotes the number of frames, and $u_t$ and $v_t$ denote bounding boxes at frame $t$ of object instances $u$ and $v$, respectively. $\mathrm{IoU}(u_t,v_t)$ returns an IoU value from $u_t$ and $v_t$ if both are detected at frame $t$ and 0 otherwise.

From the similarity matrix $S_{nm}$, which contains similarities among all combinations of object instances between ST-graphs $n$ and $m$, we find the best association $\mathcal{M}_{nm}$ that maximizes the total similarities. This problem is categorized in assignment problems. We solve it by the Hungarian algorithm \cite{kuhn1955hungarian} and obtain $\mathcal{M}_{nm}$ and $\mathcal{O}_{nm}$, where $\mathcal{O}_{nm}$ denotes a set of object instances that no associated partners can be found.

Finally, the SOIA distance, which is a distance between ST-graphs $n$ and $m$, is defined as follows:
\begin{equation}
\label{eq:st_graph_distance}
    d_{nm}=\sum_{(u,v)\in\mathcal{M}_{nm}}\frac{1}{T}\sum_{t=1}^T\mathrm{W}(u_t,v_t) + \sum_{u\in\mathcal{O}_{nm}}\frac{1}{T}\sum_{t=1}^Tm_{u_t},
\end{equation}
where
$\mathrm{W}(u_t,v_t)=\left(1-\mathrm{IoU}(u_t,v_t)\right)m_{u_tv_t}.$
Here, $m_{u_t}$ denotes the area of bounding box $u_t$, and $m_{u_tv_t}=m_{u_t}$ if $m_{u_t} > m_{v_t}$ and $m_{u_tv_t}=m_{v_t}$ otherwise.
The first term on the right side in Eq. \ref{eq:st_graph_distance} indicates the distance based on mIoU weighted by the bounding box area. The weight is necessary to prevent objects with small bounding box areas from having a large effect on the final distance. The second term largely affects the distance when large objects remain without an associated partner.

\noindent
\textbf{Selection of Positive and Negative Samples.}\quad
SOIA enables us to obtain video-to-video distance $d_{nm}$, $\forall n, m\in\mathcal{B}$, where $\mathcal{B}$ denotes a set of indices of videos in a batch.
Instead of augmenting the anchor sample in the usual CL, we select a positive sample as the closest sample to the anchor in a batch in terms of the SOIA distance.
Here, a batch often contains multiple similar scenes to the anchor, which may interfere the results of the representation learning.
To deal with this problem, we set a margin between the positive sample and the negative samples to be selected.
More specifically, for anchor sample $n^*$, we choose a candidate set of positive and negative samples: $\mathcal{D}_{n^*}=\{d_{nm}|n=n^*, m\in\mathcal{B}, m\neq n^* \}$.
This set, $\mathcal{D}_{n^*}$, is sorted in ascending order and the smallest is selected as the positive sample.
Then, a margin of $\alpha |\mathcal{B}|$ is provided and the remaining $|\mathcal{B}|-1-\alpha |\mathcal{B}|$ samples are selected as negative samples.

\subsubsection{Semi-Supervised Contrastive Learning (SSCL)}
\label{sec:semi_supervised_contrastive_learning}

In the standard self-supervised CL, positive and negative samples are determined taking no notice of their labels, as explained in Sec. \ref{sec:graph_contrastive_learning} and \ref{sec:soia}.
In addition to these samples, positive and negative samples can be defined based on the labels.
More specifically, samples with the same label as that of the anchor sample are selected as positive samples, while those with different labels are selected as negative.
CL using these two types of positive and negative samples is called the supervised CL \cite{NEURIPS2020_d89a66c7}.
Recently, Cui et al. \cite{Cui_2021_ICCV} proposed parametric contrastive learning as an extension of the original supervised CL.
The original supervised CL suffers from the class imbalance problem.
For a batch in relatively small size, the scarcity of the images of the same label may cause poor optimization.
This problem is more serious in the current semi-supervised setting because the batch also contains unlabeled data.
The parametric CL addresses this problem by introducing learnable class-wise prototypes.
As described in detail below, the proposed method in this article adopts the parametric CL with a suitable extension to the semi-supervised setting.

\noindent
\textbf{Loss Function.}\quad
Let us denote a set of negative samples corresponding to an anchor sample $n$ as $\mathcal{N}_n$ and a positive sample as $p_n$.
The positive sample $p_n$ is either generated by data augmentation (Sec. \ref{sec:graph_contrastive_learning}) or found as the closest sample in a batch using SOIA (Sec. \ref{sec:soia}).
The loss function is defined as follows:
\begin{equation}
    \mathcal{L} = \sum_{n=1}^{N} 
    \mathcal{L}_n=- \sum_{n=1}^{N}\sum_{\boldsymbol{z}_+\in \mathcal{P}_n}\log\left(
    \frac{e^{\boldsymbol{z}_+\cdot \boldsymbol{z}_n}}
    {\sum_{\boldsymbol{z}_k\in\mathcal{A}_n} e^{\boldsymbol{z}_k\cdot \boldsymbol{z}_n}}
    \right),
\label{eq:loss}
\end{equation}
where
\begin{align}
    \mathcal{P}_n &= \left\{
\begin{array}{ll}
\{\boldsymbol{z}_{p_n}, \boldsymbol{c}_{\sigma_n}\} & (\mathrm{if}\ n\ \mathrm{is}\ \mathrm{a}\ \mathrm{labeled}\ \mathrm{data}),\\
\{\boldsymbol{z}_{p_n}\} & (\mathrm{if}\ n\ \mathrm{is}\ \mathrm{an}\ \mathrm{unlabeled}\ \mathrm{data}),
\end{array}
\right.
\\
\mathcal{A}_n&=\{\boldsymbol{z}_k|k\in\mathcal{N}_n \}
\cup \{\boldsymbol{z}_{p_n}\}
\cup \mathcal{C}.
\end{align}
Here, $\mathcal{C}=\{\boldsymbol{c}_1,\dots,\boldsymbol{c}_C\}$ denotes a set of learnable class-wise prototypes, and $C$ denotes the number of scene-level classes. 
$\boldsymbol{z}_{n}$ and $\sigma_n$ denote a feature vector output from the GCN and a class label of $n$, respectively.
If $n$ is an unlabeled video, the feature vector $\boldsymbol{z}_{n}$ is optimized to be closer to only the corresponding positive sample feature vector $\boldsymbol{z}_{p_n}$ and away from the corresponding negative samples and all prototypes, which
indicates that we assume that unlabeled data is not classified into any of the predefined classes.
In our experiments, all videos that are not assigned to any of the predefined classes are treated as unlabeled videos.

\section{Evaluation}
\subsection{Evaluation Setup}

\noindent
\textbf{Datasets.}\quad
The performance of the proposed method was evaluated using Honda Research Institute Driving Dataset (HDD) \cite{8578901}.
The HDD includes 104 hours egocentric videos of real human driving in the San Francisco Bay Area, and frame-level annotations of vehicle actions.
The videos have a resolution of $1280\times 720$ at 30fps.
Following prior works \cite{8578901,Li2020Learning3E}, we used labels of 11 Goal-oriented actions (e.g., left turn, right lane change, merge, etc.) and 6 Causes (e.g., congestion, sign, red light, etc.).
The 6 Causes include the five different stopping actions depending on its cause and one deviating action.
In addition, following a prior work \cite{8578901}, we split 137 sessions of the HDD into 100 sessions for training and 37 sessions for testing.
Note that in order to keep the experimental conditions the same as in the prior work \cite{8578901}, the model is trained with 17 labels (11 Goal-oriented actions + 6 Causes) only when evaluating the 6 Causes.

\noindent
\textbf{Implementation Details.}\quad
As the object detector and tracker, we adopted Faster-RCNN \cite{NIPS2015_14bfa6bb} and QDTrack \cite{Pang2021QuasiDenseSL}, respectively.
The trained model with the BDD100k dataset \cite{Yu2020BDD100KAD} is distributed by the authors of \cite{Pang2021QuasiDenseSL}, which we used to construct ST-graphs.
The object tracking was performed on the HDD of 30fps, and then the tracking results were downsampled to 2.5fps.
To construct ST-graphs for unlabeled video clips, the original video was clipped every 4 seconds at equally interval. The interval was determined to adequately cover individual actions to be detected.
The selected optimizer was Adam \cite{Kingma2015AdamAM} with default parameters. The initial learning rate was 0.01, and the cosine annealing learning rate scheduler was employed.
All experiments were performed on a workstation equipped with Tesla V100 GPUs.

\noindent
\textbf{Evaluation Metrics.}\quad
We used mean average precision (mAP) to evaluate the performance of the proposed method. The AP is the area under the precision-recall curve, which is used as standard in the prior works.

\noindent
\textbf{Baselines.}\quad
The proposed method is compared with existing methods of baselines.
These methods are not limited to ego-vehicle action recognition, but include general action recognition methods such as C3D \cite{Tran2015LearningSF} and I3D \cite{Carreira2017QuoVA}.
Following \cite{8578901}, we classify the existing methods into in online and offline settings.
In the online setting, they infer an action label of the ego-vehicle each time a new frame is entered.
In the offline setting, on the other hand, an action label is inferred for each short video clip.
The proposed method falls into the latter category.

\begin{table*}[hbtp]
  \centering
  \resizebox{\textwidth}{!}{\begin{tabular}{cccccccccccccccc}
    \hline
    & & & \multicolumn{11}{c}{Individual actions} \\
    \cline{4-13}
    \multirow{2}{*}{Methods} & Online & Train & intersection & &  & L lane & R lane & L lane & R lane & crosswalk & railroad & & & & Overall  \\
     & /Offline & Data & passing & L turn & R turn & change & change & branch & branch & passing & passing & merge & u-turn &  & mAP   \\
    \hline
  CNN \cite{8578901} & \multirow{5}{*}{Online} & L & 53.4 & 47.3 & 39.4  & 23.8 & 17.9 & 25.2 & 2.9 & 4.8 & 1.6 & 4.3 & 7.2 &  & 20.7 \\
  CNN-LSTM \cite{8578901} & & L &65.7  & 57.7 & 54.4  & 27.8 & 26.1 & 25.7 & 1.7 & 16.0 & 2.5 & 4.8 & 13.6 &  & 26.9 \\
  ED \cite{9009797} & & L & 63.1 & 54.2 & 55.1 & 28.3 & 35.9 & 27.6 & 8.5 & 7.1 & 0.3 & 4.2 & 14.6 & & 27.2  \\
  TRN \cite{9009797} & & L & 63.5 & 57.0 & 57.3 & 28.4 & 37.8 & 31.8 & 10.5 & 11.0 & 0.5 & 6.3 & 16.7 & & 33.7 \\
  DEPSEG-LSTM \cite{narayanan2018semi} & & L & 70.9 & 63.4 & 63.6 & 48.0 & 40.9 & 39.7 & 4.4 & 16.1 & 0.5 & 6.3 & 16.7 & & 33.7  \\
  C3D \cite{Tran2015LearningSF} & & L & 72.8 & 64.8 & 71.7 & 53.4 & 44.7 & 52.2 & 3.1 & 14.6 & 2.9 & 10.6 & 15.8 & & 37.0 \\
  \hline
  C3D \cite{Tran2015LearningSF} & \multirow{4}{*}{Offline} & L & 82.4 & 77.4 & 80.7 & 67.9 & 56.9 & 59.7 & 5.2 & 17.4 & 3.9 & 20.1 & 29.5 & & 45.5 \\
  I3D \cite{Carreira2017QuoVA} & & L & 85.6 & 79.1 & 78.9 & 74.0 & 62.4 & 59.0 & 14.3 & 29.8 & 0.1 & 20.1 & 41.4 & & 49.5 \\
  GCN \cite{Li2020Learning3E} & & L & 85.5 & 77.9 & 79.1 & \textbf{76.0} & 62.0 & \textbf{64.0} & 19.8 & 29.6 & 1.0 & 27.7 & 39.9 & & 51.1  \\
  Ours (SCL) & & L+U & 98.3 & \textbf{94.1} & \textbf{95.8} & 62.6 & 67.3 & 53.4 & 28.4 & \textbf{78.0} & 1.2 & 22.2 & \textbf{60.0} &  & 60.1\\
  Ours (GCL) & & L+U &\textbf{98.4} & 93.9 & 95.5 & 64.2 & \textbf{69.0} & 55.8 & \textbf{34.5} & 73.4 & \textbf{24.4} & \textbf{42.4} & 30.0 &  & \textbf{62.0}\\
    \hline
  \end{tabular}}
\caption{Comparison of classification performance between the proposed and existing methods for labels of 11 Goal-oriented actions. The column for Train Data shows datasets used for training (L=labeled data, U=unlabeled data).}
\label{table:main_result}
\end{table*}

\begin{table*}[hbtp]
  \centering
  \resizebox{\textwidth}{!}{\begin{tabular}{cccccccccc}
    \hline
    & \multicolumn{6}{c}{Individual actions} \\
    \cline{3-7}
    \multirow{2}{*}{Methods} & Train & Stop for & Stop for & Stop for & Stop for & Deviate for & Stop for & Overall  \\
     & Data &  Congestion & Sign & Red Light & Crossing Vehicle & Parked Car & Crossing Pedestrian  & mAP   \\
    \hline
  I3D \cite{Carreira2017QuoVA} & L & 64.8 & 71.7 & 63.6 & 21.5 & 15.8 & 26.2 & 43.9 \\
  GCN \cite{Li2020Learning3E} & L & 74.1 & 72.4 & 76.3 & 26.9 & 20.4 & 29.0 & 49.9  \\
  Ours (SCL) & L+U & 95.8 & \textbf{92.2} & 76.9 & \textbf{67.3} & \textbf{61.9} & 65.7 & \textbf{76.6} & \\
  Ours (GCL) & L+U & \textbf{95.9} & 90.9 & \textbf{81.0} & 54.7 & 53.7 & \textbf{69.0} & 74.1 & \\
    \hline
  \end{tabular}}
\caption{Comparison of classification performance between the proposed and existing methods for labels of 6 Causes.}
\label{table:main_result_cause}
\end{table*}

\subsection{Evaluation Results for Labeled Videos}
Table \ref{table:main_result} and \ref{table:main_result_cause} present the experimental results obtained by applying the proposed method to the HDD.
The tables show the results of the proposed method based on two learning methods: SOIA-based CL (SCL) and graph contrastive learning (GCL).
Table \ref{table:main_result} shows the results of the 11 Goal-oriented actions.
As can be seen, the performances of the two proposed methods overcome those of the existing methods.
In particular, the proposed methods achieve outstanding performance for individual actions that are characterized primarily by relationships among object instances, e.g., intersection passing, L/R turn, U-turn.
On the other hand, individual actions such as L/R lane change, L/R lane branch and merge are strongly affected by other environmental conditions such as lane lines.
Even for these actions, the performances of the proposed methods are competitive with those of the existing methods.
The railroad passing is the most difficult action to predict because information about the background of the scene is more important than the action of the object instances.

Table \ref{table:main_result_cause} presents the results of the 6 Causes.
As can be seen, the proposed method significantly improves the performance compared to the existing methods for the most labels in the 6 Causes.
In fact, scenes with these labels are strongly affected by relationships between among object instances.
For example, in the ``Stop for Congestion'' scene, a car is stopped in front of the ego-vehicle, and in the ``Stop for Red Light'' scene, the ego-vehicle is at an intersection and there are no cars in from of the ego-vehicle. These results might imply that
the semi-supervised learning using relationships among object instances is more effective and leads to the performance improvement.

Furthermore, there are also distinctive differences between SCL and GCL.
Since SCL uses the SOIA distances, which are strictly defined from relationships between object instances, the performance improvement is more significant for actions that better fit the assumption (e.g., U-turn and Stop for Crossing Vehicle).
On the other hand, for the other individual actions that are strongly affected by other environmental conditions (e.g., railroad passing and merge), the benefits of the SOIA distances are limited, while GCL, which does not use such strict distances, outperforms SCL.

\subsubsection{Difference between learning methods}

\begin{figure}[htbp!]
    \hspace{-1.05cm}
    \begin{tabular}{c}
    \begin{minipage}[t]{0.49\linewidth}
        \centering
      \includegraphics[clip, width=10.0cm]{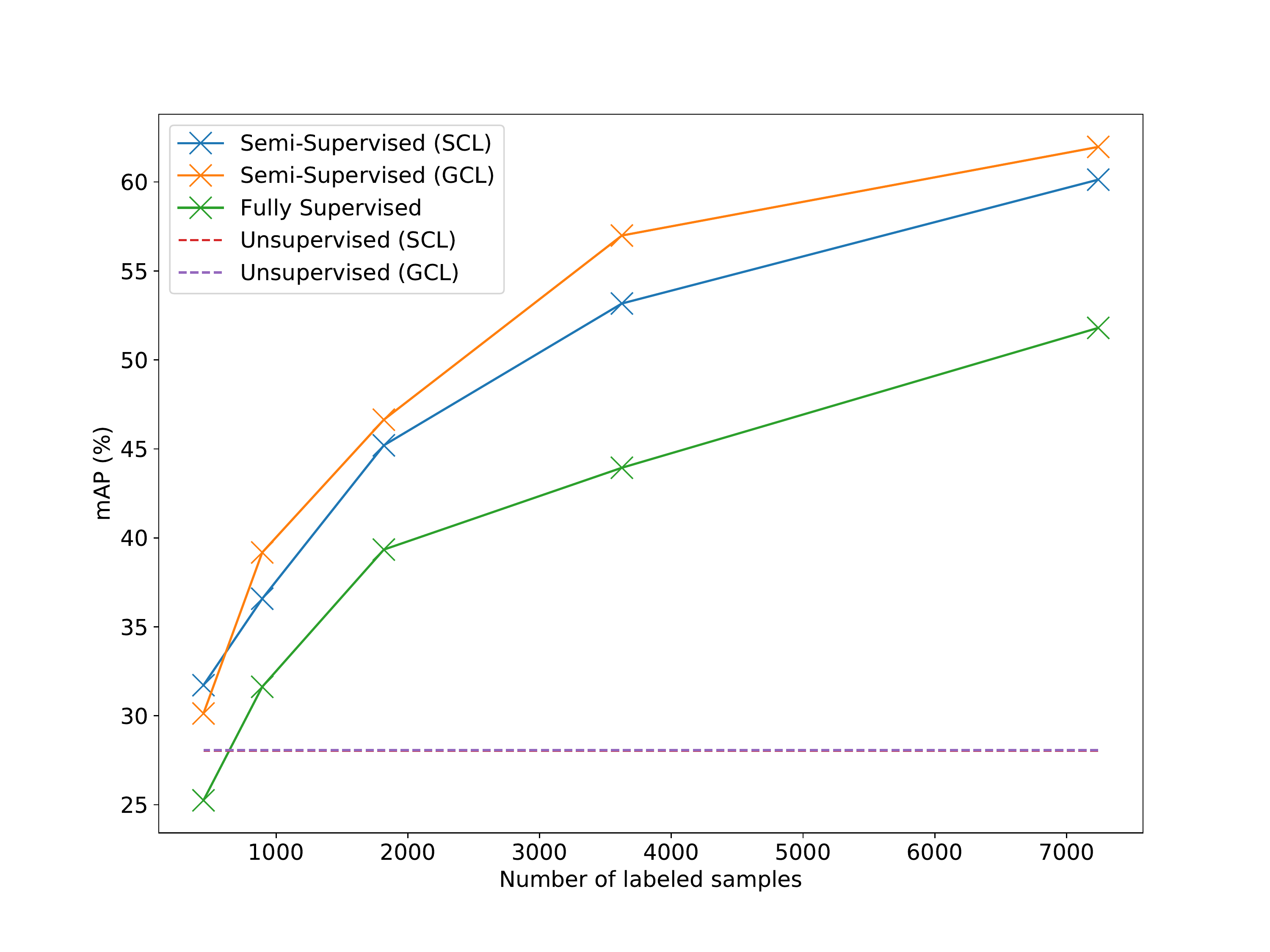}
      \end{minipage}
    \end{tabular}
    \caption{Comparison of classification performance between the different learning methods in the proposed method when varying the number of labeled samples.}
    \label{fig:semi_sp_un_comp}
\end{figure}

We compare the results of three learning methods: fully-supervised, semi-supervised and unsupervised learning.
In the three settings, we used the same GCN architecture.
In the fully-supervised setting, we trained the model using only labeled data.
When no unlabeled data is used, the loss function in Eq. \ref{eq:loss} reduces to the usual cross entropy loss.
In the unsupervised setting, we trained the model without using labeled data and used the same loss function in Eq. \ref{eq:loss}.

Figure \ref{fig:semi_sp_un_comp} shows the results of the performance comparison between the three settings when varying the number of labeled samples.
The used labeled data are labels of 11 Goal-oriented actions.
As can be seen, the proposed method in the semi-supervised setting outperforms that in the supervised setting.
Further, although their performances naturally degrade as the number of labeled samples decreases, the performance in the semi-supervised setting never becomes lower than that in the unsupervised setting. On the other hand, the performance in the supervised setting is lower than that in the unsupervised setting when the number of labeled samples is small in Fig. \ref{fig:semi_sp_un_comp}.
These are because that as the number of labeled data decreases, the semi-supervised setting reduces to the unsupervised setting while the fully-supervised setting reduces to the setting of random guess.

\begin{figure*}[h]
    \begin{tabular}{c}
      \begin{minipage}[t]{0.95\linewidth}
        \centering
        \includegraphics[width=17.0cm]{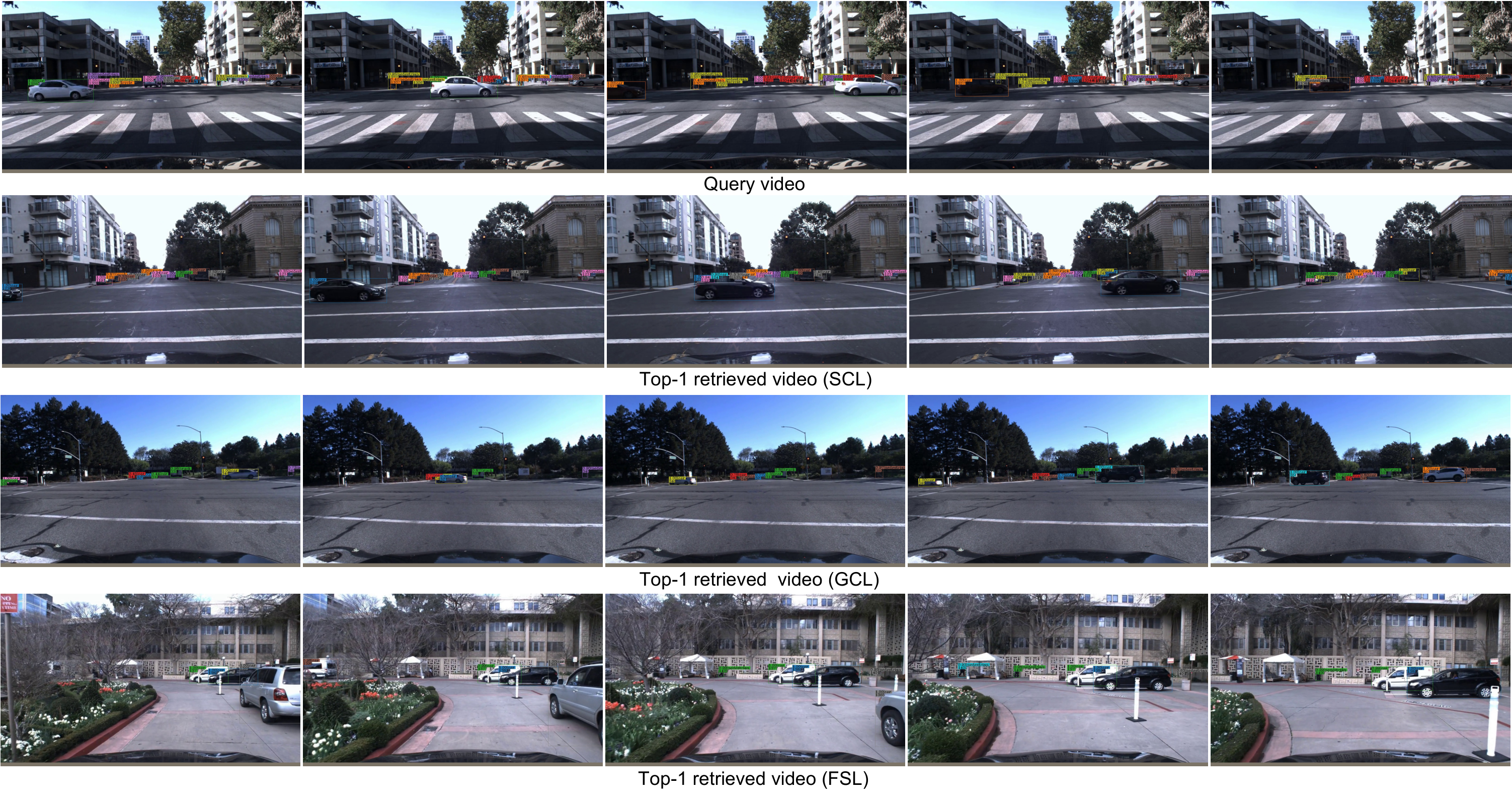}
      \end{minipage}
    \end{tabular}
    \caption{Five frames extracted at equal interval from query and retrieved videos. The top row shows a query video, and the second, third and forth rows show top-1 retrieved videos obtained from the proposed methods with SCL, GCL and FSL, respectively.}
    \label{fig:unlabeled_scenes}
  \end{figure*}

% \begin{figure*}[h]
%     \begin{tabular}{cc}
%       \begin{minipage}[t]{0.95\linewidth}
%         \centering
%         \includegraphics[width=17.0cm]{figures/query.pdf}
%       \end{minipage} \\
%       \begin{minipage}[t]{0.95\linewidth}
%         \centering
%         \includegraphics[width=17.0cm]{figures/scl.pdf}
%       \end{minipage} \\
%       \begin{minipage}[t]{0.95\linewidth}
%         \centering
%         \includegraphics[width=17.0cm]{figures/gcl.pdf}
%       \end{minipage} \\
%       \begin{minipage}[t]{0.95\linewidth}
%         \centering
%         \includegraphics[width=17.0cm]{figures/sp.pdf}
%       \end{minipage}
%     \end{tabular}
%     \caption{Five frames extracted at equal interval from query and retrieved videos. The top row shows a query video, and the second, third and forth rows show top-1 retrieved videos obtained from the proposed methods with SCL, GCL and FSL, respectively.}
%     \label{fig:unlabeled_scenes}
%   \end{figure*}

\subsection{Evaluation Results for Unlabeled Videos}
In this subsection, we qualitatively evaluate the video-to-video distances learned by the proposed methods.
We provide the comparison results between three learning methods: SCL, GCL and fully-supervised learning (FSL).
Here, because of the difficulty of quantitatively evaluating the quality of the distances between unlabeled videos, we instead present as many query-retrieval examples as possible to provide a qualitative evaluation.
Specifically, we chose a query video from unlabeled videos in the validation set and searched the nearest neighbor video of the query video in the embedding space learned by each method.
The nearest neighbor video was found from all videos including both labeled and unlabeled videos in the train set.
The distances were measured using cosine similarities between feature vectors output from GCNs.
For reasons of space limitation, only one example can be presented here; the rest are presented in Appendix C.

Figure \ref{fig:unlabeled_scenes} shows a query-retrieval example obtained from the three proposed methods with different learning methods: SCL, GCL and FSL.
These methods were trained with labels of the 11 Goal-oriented actions.
Five frames were extracted at equal intervals from an unlabeled 4-second video.
The five images in the top row are a query video, and those in the second, third and forth rows are top-1 retrieved videos obtained from the proposed methods with SCL, GCL and FSL, respectively.
In the query video, the ego-vehicle is stopped at an intersection and a white car is crossing in front of the ego-vehicle from left to right.
In the video retrieved by the proposed method with SCL (second row), we can see that the ego-vehicle stopped at an intersection and a car crossing in front of the ego-vehicle from left to right, which is similar to the scene in the query video.
In the video retrieved by the proposed method with GCL (third row), we can also see that the ego-vehicle stopped at an intersection.
However, the behaviors of the other cars are quite different from those in the query video (e.g., cars are crossing from right to left).
Finally, in the video retrieved with using FSL, the ego-vehicle is not at an intersection, and the video is completely different from the query video.
From the above, it can be said that, at least in this example, the video-to-video distance learned with SCL is the most sensible.

A similar tendency was observed in other retrieved results (see Appendix C for more samples). FSL results are often far from scenes in the query videos. We believe this is because FSL only considers predefined labeled videos. In fact, GCL, which also considers unlabeled videos, could provide more sensible distances than FSL. However, GCL considers only distances based on graph structures, which are not necessarily in accord with sensible distances between driving videos. On the other hand, SCL explicitly considers distances based on movements of object instances (SOIA distance). Therefore, they can provide more sensible distances.
Note that the labeled data used for training in this experiment consist of only the 11 Goal-oriented actions and do not include labels related to the scene of the query video, such as ``Stop for Red Light'' and ``Stop for Crossing Vehicle''.

\section{Limitation}
As mentioned in Sec. \ref{sec:semi_supervised_contrastive_learning}, we assume that the unlabeled data are not classified into any of predefined classes. This is required to be confirmed by human annotators.
In many cases, however, these unlabeled data are collected in large quantities during the annotation process, because the majority of the collected videos are irrelevant scenes to the predefined classes.

\section{Conclusions}
We proposed a method that can set proper and sensible distances between all of both labeled and unlabeled driving videos.
The proposed method is based on semi-supervised contrastive learning (SSCL).
In the SSCL, the GCN was trained using both distances determined by the labels and distances without the labels.
These distances were used to generate positive and negative samples in a batch.
We investigated two kinds of CL. In addition to the standard GCL, we proposed SOIA-based CL.
In the evaluation for unlabeled videos, we observed that the distance learned by SOIA-based CL was the most proper and sensible.

Furthermore, we quantitatively evaluate the proposed method in terms of the classification accuracy in the ego-vehicle action recognition.
We presented the results of three different evaluations in which the proposed method was trained with each of SCL, GCL, and FSL.
The results show that the proposed method with SCL and GCL achieved the state-of-the-art performance on HDD.
In addition, the proposed method with SCL and GCL outperforms that with FSL.
Therefore, we found that using unlabeled videos for training is effective in improving the classification performance of the ego-vehicle action recognition.

\clearpage

{\small
\bibliographystyle{ieee_fullname}
\bibliography{egbib}
}

\clearpage
\appendix

\section{GCN architecture}
\label{sec:gcn_arc}
The constructed ST-graphs $G_n$, $\forall n\in\{1,\dots,N\}$, where $N$ denotes the number of video clips, are fed into a GCN.
Following \cite{pmlr-v97-li19d}, our GCN model consists of three parts: an encoder, propagation layers, and an aggregator.

\noindent
\textbf{Encoder.}\quad
Node attributes $\boldsymbol{s}_i$ and $\boldsymbol{g}_i$ are separately fed into multilayer perceptrons (MLPs) first:
\begin{align}
    \boldsymbol{s}^{\prime}_i=\mathrm{MLP}_{s}(\boldsymbol{s}_i),
    \quad\forall i\in V_n \\
    \boldsymbol{g}^{\prime}_i=\mathrm{MLP}_{g}(\boldsymbol{g}_i),
    \quad\forall i\in V_n.
\end{align}
Here $\boldsymbol{s}^{\prime}_i\in\mathbb{R}^{32}$ and $\boldsymbol{g}^{\prime}_i\in\mathbb{R}^{32}$ are the same 32-dimensional.
$\boldsymbol{s}_i$ and $\boldsymbol{g}_i$ can have different properties due to the one-hot encoding of $\boldsymbol{s}_i$. Therefore, it is useful to first map the node attributes at the encoder, rather than feeding them directly to the propagation layer.
The encoded attributes are concatenated as $\boldsymbol{x}^{(0)}_i=[\boldsymbol{g}^{\prime}_i,\boldsymbol{s}^{\prime}_i]$.

\noindent
\textbf{Propagation Layers.}\quad
In a propagation layer, the features of each node are aggregated according to adjacencies defined by the ST-graphs.
Our GCN model adopts the local extrema convolution (LEConv) \cite{Ranjan2020ASAPAS}, whose update formula for the $l$th layer is defined as follows:
\begin{equation}
    \boldsymbol{x}^{(l+1)}_i = 
    \sigma\left(
    \boldsymbol{x}^{(l)}_i W_1^{(l)}
    + \sum_{j\in\partial i} e_{ij} \left(\boldsymbol{x}^{(l)}_i W_2^{(l)}-\boldsymbol{x}^{(l)}_j W_3^{(l)}\right)
    \right),
\end{equation}
for $\forall i\in V_n$, where $\partial i$ denotes a set of indices of adjacent nodes of $i$, and $W_1^{(l)}$, $W_2^{(l)}$ and $W_3^{(l)}$ denote learnable parameters. $\sigma(\cdot)$ denotes the ReLU activate function.

\noindent
\textbf{Aggregator.}\quad
As outputs of the propagation layers, we obtain node representations $\boldsymbol{x}_i^{(L)}$, $\forall i\in V_n$, where $L$ denotes the number of propagation layers ($L=3$ in our setting).
The aggregator performs a pooling operation to output a graph-level representation.
In our GCN model, in order to explicitly learn instance-level features, we introduce an instance-level pooling.
As a result, graph-level representation $\boldsymbol{z}_n$ can be obtained as the output of the following aggregator operation:
\begin{equation}
    \boldsymbol{z}_n=\mathrm{MLP}_{\mathrm{1}}\left(\sum_{u\in\mathcal{I}_n}\mathrm{MLP}_{\mathrm{2}}\left(\sum_{i\in\mathcal{N}_{nu}}\mathrm{MLP}_{\mathrm{3}}(\boldsymbol{x}_i^{(L)})\right)\right),
\end{equation}
$\forall n\in\{1,\dots,N\}$, where $\mathcal{N}_{nu}$ denotes a set of nodes corresponding to object instance $u$ in ST-graph $n$, and $\mathcal{I}_n$ denotes a set of object instances in graph $n$. 

\section{Ablation Studies}
To provide a further understanding of the proposed method, we perform two kinds of ablation studies. First, we investigate the effect of removing each of the three types of node attributes: semantic labels, geometric features of bounding boxes, and interaction with lane lines on classification performance.
Second, we examine how effective on the classification performance by alleviating the significant imbalance between the number of labeled and unlabeled videos.

\begin{table*}[hbtp]
  \centering
  \resizebox{\textwidth}{!}{\begin{tabular}{cccccccccccccc}
    \hline
    & & \multicolumn{11}{c}{Individual actions} \\
    \cline{2-13}
    \multirow{2}{*}{Methods} & intersection & &  & L lane & R lane & L lane & R lane & crosswalk & railroad & & & & Overall  \\
     &  passing & L turn & R turn & change & change & branch & branch & passing & passing & merge & u-turn &  & mAP   \\
     \hline
  SCL & 98.3 & 94.1 & 95.8 & 62.6 & 67.3 & 53.4 & 28.4 & 78.0 & 1.2 & 22.2 & 60.0 &  & 60.1\\
  SCL (w/o semantic) & 97.9 & 94.8 & 95.8 & 60.5 & 57.3 & 53.8 & 20.6 & 77.8 & 3.5 & 28.7 & 55.1 &  & 58.7 \\
  SCL (w/o bbox) & 98.2 & 92.5 & 94.5 & 55.0 & 53.0 & 55.2 & 22.9 & 73.2 & 1.2 & 23.5 & 30.2 &  & 54.5 \\
  SCL (w/o lane) & 95.8 & 92.8 & 93.5 & 50.3 & 45.6 & 25.2 & 13.0 & 58.1 & 1.2 & 10.9 & 52.0 &  & 48.9\\
    \hline
  \end{tabular}}
\caption{Comparison of classification performance of SCL when removing each of the three kinds of node attributes. The top row shows the results with three kinds of node attributes. The second, third and forth rows show the results without using semantic labels (w/o semantic), geometric features of bounding boxes (w/o bbox) and interaction with lane lines (w/o lane), respectively. }
\label{table:wo_lane_line}
\end{table*}

\noindent
\subsection{Effect of node attributes.}
\label{sec:ablation_node_attribute}
Table \ref{table:wo_lane_line} shows the results of ablation studies for node attributes.
These results correspond to the classification performance of SCL for labels of 11 Goal-oriented actions.
The top row in table \ref{table:wo_lane_line} shows the results with three kinds of node attributes.
The second, third and forth rows show the results without using semantic labels, geometric features of bounding boxes (bbox features) and interaction with lane lines (lane line features), respectively.

Without using the semantic labels, the mAP values are not significantly different with those of SCL with all kinds of node attributes.
This indicates that distinguishing between kinds of object instances has little impact on the classification performance of the goad-oriented actions.
However, when recognizing other kinds of scenes including unlabeled scenes, semantic labels in node attribute would be possible to play an importance role.
For example, Fig. \ref{fig:example_intersection} shows pedestrians crossing a crosswalk. To recognize this scenes, it is important to distinguish pedestrians from vehicles.

When the box features are not taken into account, the mAP values largely drop.
We consider that this is because without the bounding box information, it cannot accurately track moves of object instances.
However, the performance degradation is less than the case without lane line features.
We believe that this is because relative positions between object instances, which are given from a ST-graph, can also be used to track moves of object instances.
As a result, the performance degradation may have been reduced compared to the case without lane line features.

Finally, when the lane line features were not included in node attributes, the performance degradation was greatest in the three cases.
We consider the reason for this is that the lane line features cannot be substituted for other features.
In fact, the AP values of individual actions significantly influenced by lane lines (L/R lane change, L/R lane branch and merge) significantly dropped when lane line features were not used.

\begin{figure}[htbp!]
    \hspace{-1.05cm}
    \begin{tabular}{c}
    \begin{minipage}[t]{0.49\linewidth}
        \centering
      \includegraphics[clip, width=10.0cm]{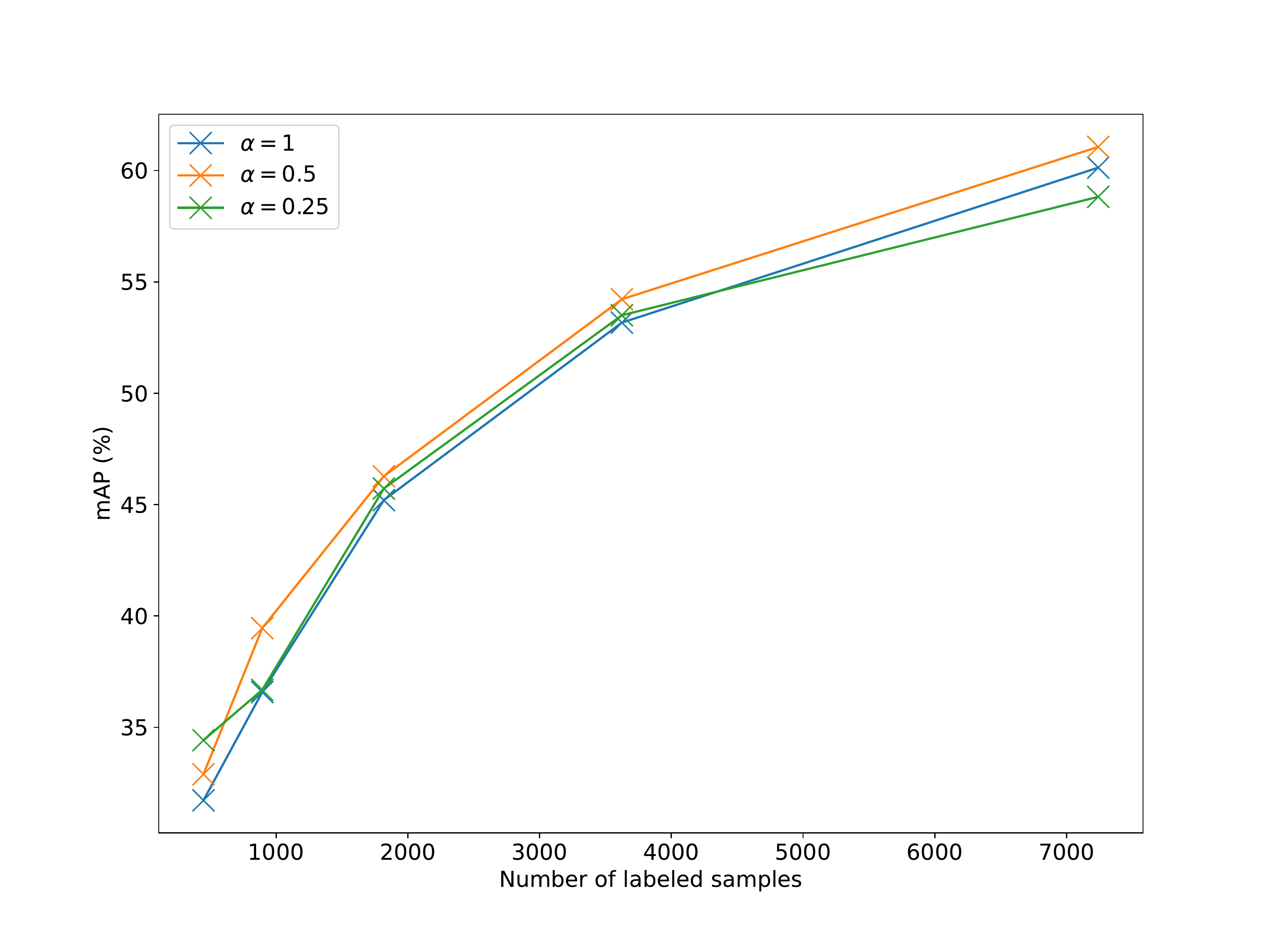}
      \end{minipage}
    \end{tabular}
    \caption{Change in classification performance of the proposed method with SCL when varying $\alpha$.}
    \label{fig:weight_comp}
\end{figure}

\noindent
\subsection{Varying weights in the loss function.}
\label{sec:ablation_weights}
Since it is much easier to collect unlabeled videos than to collect labeled videos, the number of unlabeled data is much more than that of labeled data in many cases.
If the focus is solely on improving the classification performance, this imbalance can have a negative impact on the performance.
A straightforward approach to alleviate the imbalance is to introduce weights into the loss function.
Therefore, in this section, we investigate how much the classification performance is improved when reducing the weight of unlabeled video.
The loss function with explicitly introduced weights is
\begin{equation}
    \mathcal{L} = \sum_{n=1}^{N} 
    \mathcal{L}_n=- \sum_{n=1}^{N} \alpha_n\sum_{\boldsymbol{z}_+\in \mathcal{P}_n}\log\left(
    \frac{e^{\boldsymbol{z}_+\cdot \boldsymbol{z}_n}}
    {\sum_{\boldsymbol{z}_k\in\mathcal{A}_n} e^{\boldsymbol{z}_k\cdot \boldsymbol{z}_n}}
    \right),
\label{eq:weight_loss}
\end{equation}
where
\begin{align}
\alpha_n &= \left\{
\begin{array}{ll}
1 & (\mathrm{if}\ n\ \mathrm{is}\ \mathrm{a}\ \mathrm{labeled}\ \mathrm{data}),\\
\alpha & (\mathrm{if}\ n\ \mathrm{is}\ \mathrm{an}\ \mathrm{unlabeled}\ \mathrm{data}).
\end{array}
\right.
\end{align}
Here, $\alpha$ controls the strength of the effect of unlabeled videos.
When $\alpha=1$, Eq. \ref{eq:weight_loss} is equivalent to Eq. 8 in the main text.

Figure \ref{fig:weight_comp} shows overall mAP values for labels of 11 Goal-oriented actions when $\alpha=1, 0.5$ and $0.25$.
As can be seen, although the mAP values are highest at $\alpha=0.5$ in most cases, the difference is slight.
The smaller the number of labeled videos, the larger the difference between the number of labeled and unlabeled videos.
However, even the number of labeled videos was very small, the effect of the introduction of weights was not able to be confirmed.
Therefore, in the other experiments in this paper, the value of $\alpha$ was fixed at 1.

\section{Query-Retrieval Examples of Unlabeled Videos}
\begin{figure*}[h]
    \begin{tabular}{c}
      \begin{minipage}[t]{0.95\linewidth}
        \centering
        \includegraphics[width=17.0cm]{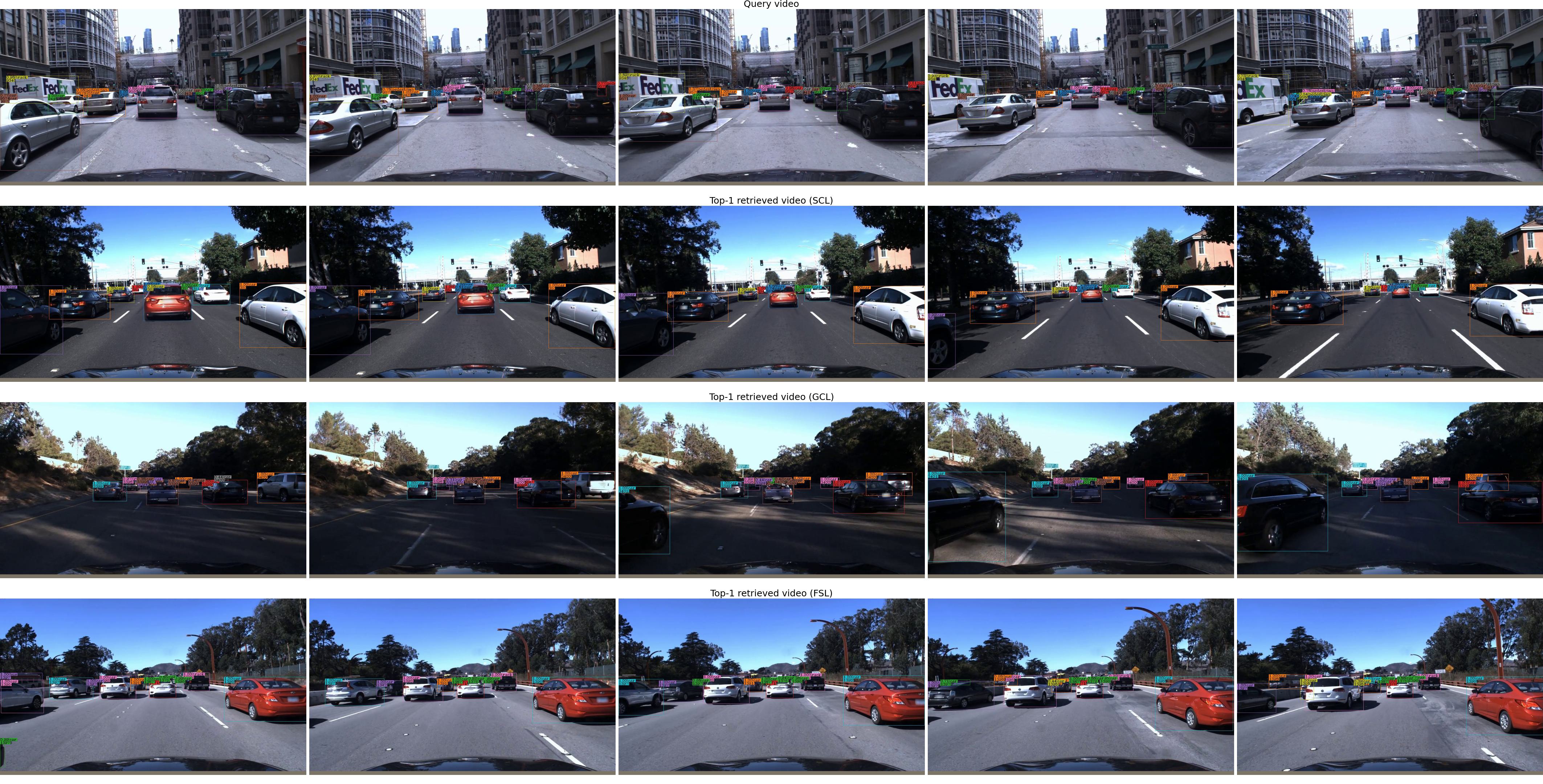}
      \end{minipage}
    \end{tabular}
    \caption{Five frames extracted at equal interval from query and retrieved videos. The top row shows a query video, and the second, third and forth rows show top-1 retrieved videos obtained from the proposed methods with SCL, GCL and FSL, respectively. In the query video, the ego-vehicle is on a busy road. The second row of video similarly shows a crowded driving scene. Note that the proposed methods primarily focus on the relationship between object instances detected in video and do not consider environmental conditions such as road conditions, surrounding buildings and nature.}
    \label{fig:sample_1}
  \end{figure*}

\begin{figure*}[h]
    \begin{tabular}{c}
      \begin{minipage}[t]{0.95\linewidth}
        \centering
        \includegraphics[width=17.0cm]{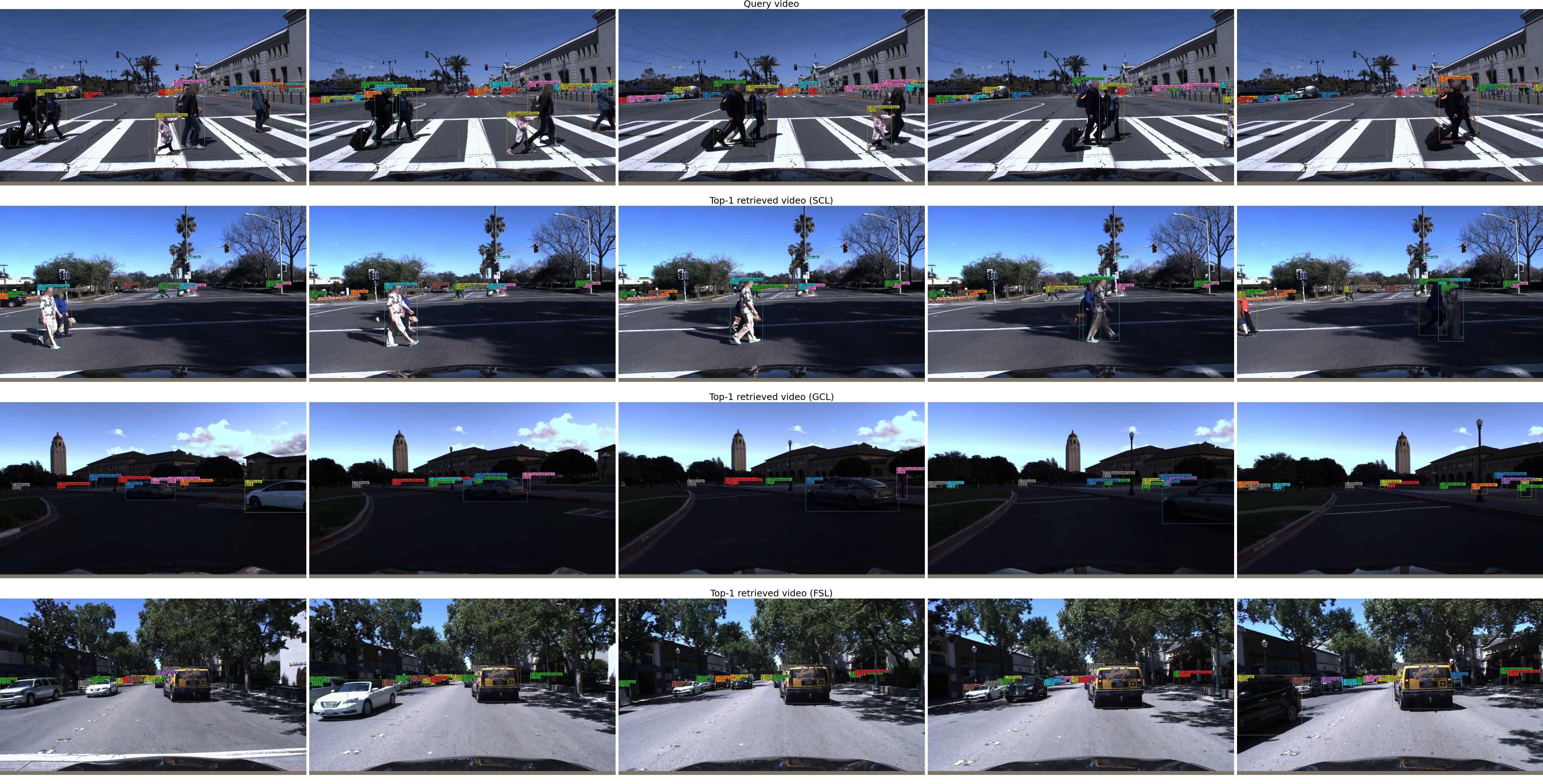}
      \end{minipage}
    \end{tabular}
    \caption{}
    \label{fig:example_intersection}
  \end{figure*}

\begin{figure*}[h]
    \begin{tabular}{c}
      \begin{minipage}[t]{0.95\linewidth}
        \centering
        \includegraphics[width=17.0cm]{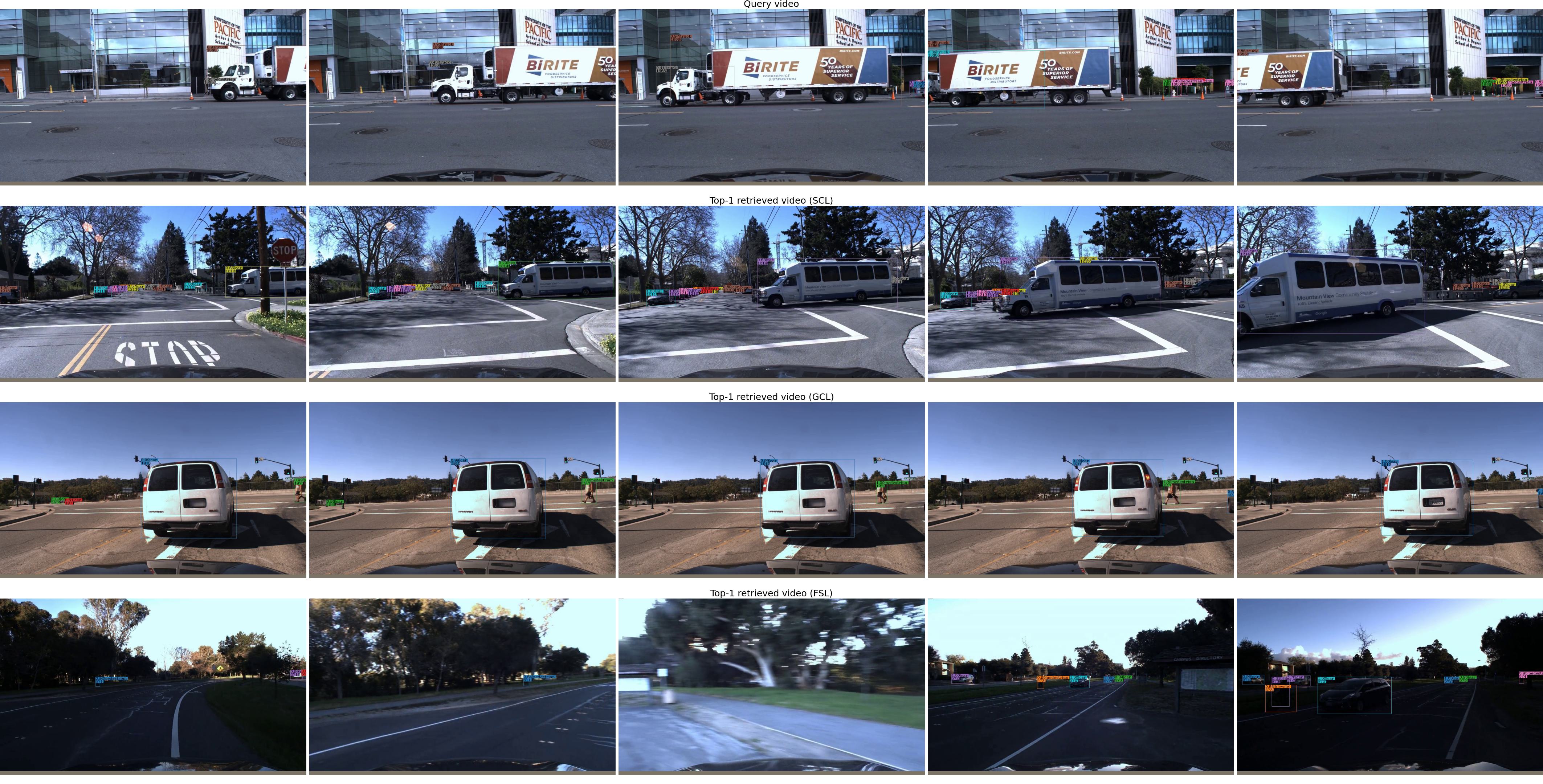}
      \end{minipage}
    \end{tabular}
    \caption{}
    \label{fig:sample_last}
  \end{figure*}

In this Appendix, we present query-retrieval examples to qualitatively evaluate video-to-video distances learned by the proposed methods. 
As described in Sec. 4.3 of the main text, we chose a query video from unlabeled videos in the validation set and searched the nearest neighbor video on the query video in the embedding space learned by each method.
The nearest neighbor video was found from all videos including both labeled and unlabeled videos in the train set.
The distances were measured using cosine similarities between feature vectors output from the GCN.
In Figs \ref{fig:sample_1}-\ref{fig:sample_last}, the remaining samples, which could not be included in the main text due to space limitations, are presented.

In addition, we present average SOIA distances between query videos and corresponding top-1 retrieved videos in Table \ref{table:soia_distance}.
As can been seen, retrieved videos by using SCL have the smallest average SOIA distance to the corresponding query videos.

\begin{table}[hbtp]
  \centering
  \begin{tabular}{cc}
    \hline
    Methods & Average SOIA distances ($\times 10^4$)\\
    \hline
  SCL & 8.68\\
  GCL & 9.83\\
  FSL  & 11.69\\
    \hline
  \end{tabular}
\caption{Average SOIA distance between query videos and corresponding top-1 retrieved videos.}
\label{table:soia_distance}
\end{table}

 \begin{figure*}[h]
    \begin{tabular}{c}
      \begin{minipage}[t]{0.95\linewidth}
        \centering
        \includegraphics[width=17.0cm]{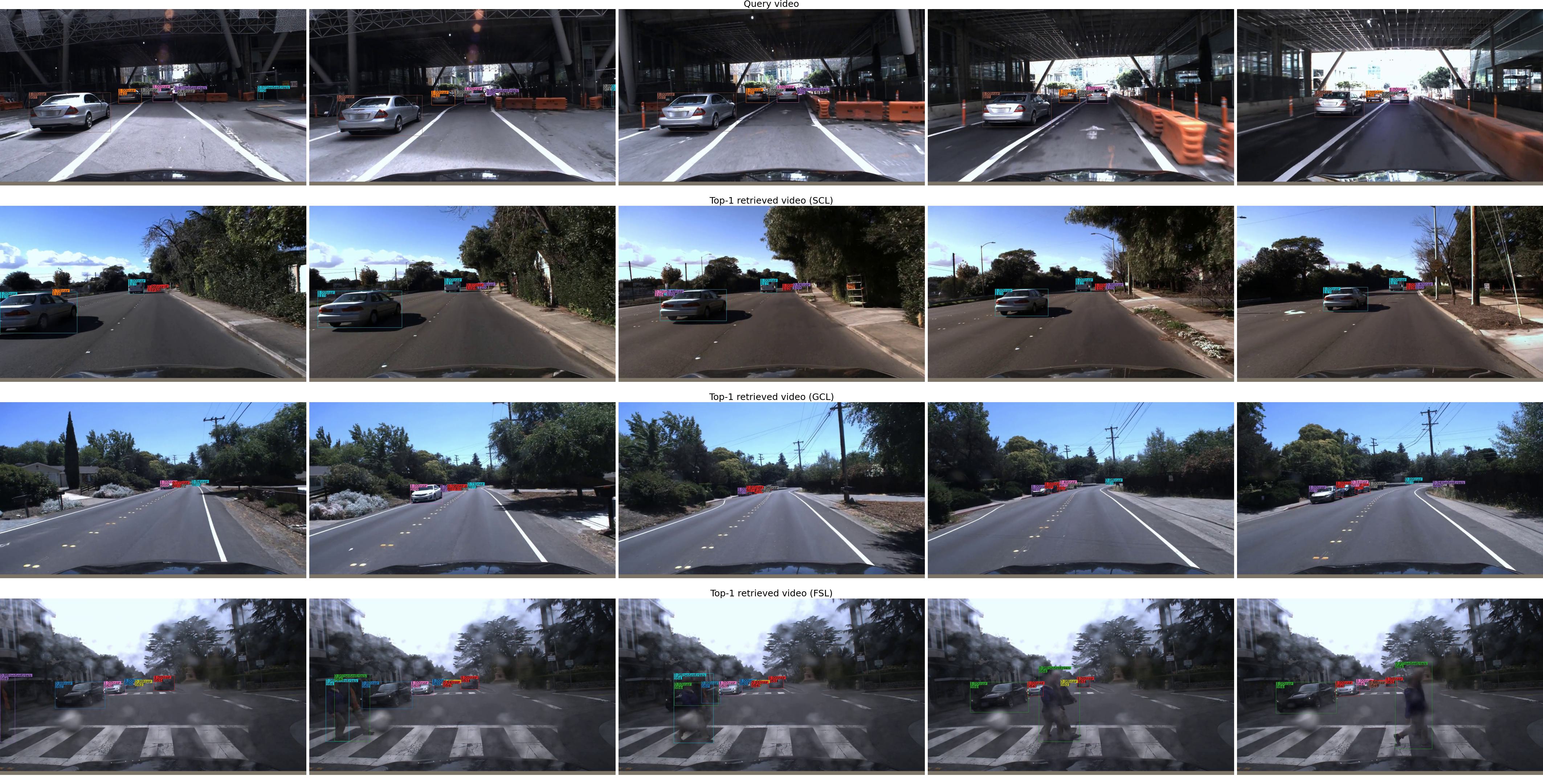}
      \end{minipage}
    \end{tabular}
    \caption{}
  \end{figure*}
  
 \begin{figure*}[h]
    \begin{tabular}{c}
      \begin{minipage}[t]{0.95\linewidth}
        \centering
        \includegraphics[width=17.0cm]{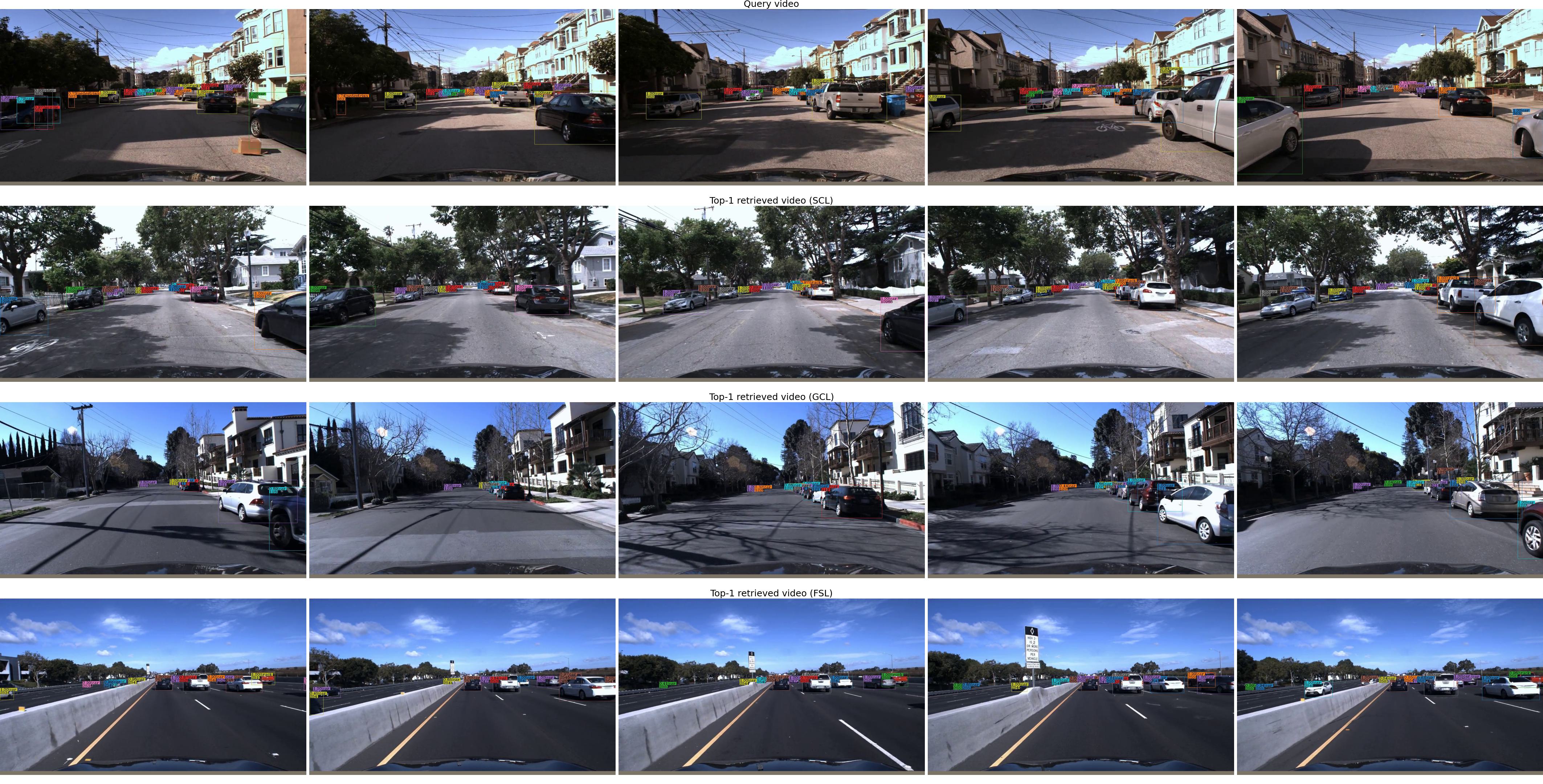}
      \end{minipage}
    \end{tabular}
    \caption{}
  \end{figure*}
  
\begin{figure*}[h]
    \begin{tabular}{c}
      \begin{minipage}[t]{0.95\linewidth}
        \centering
        \includegraphics[width=17.0cm]{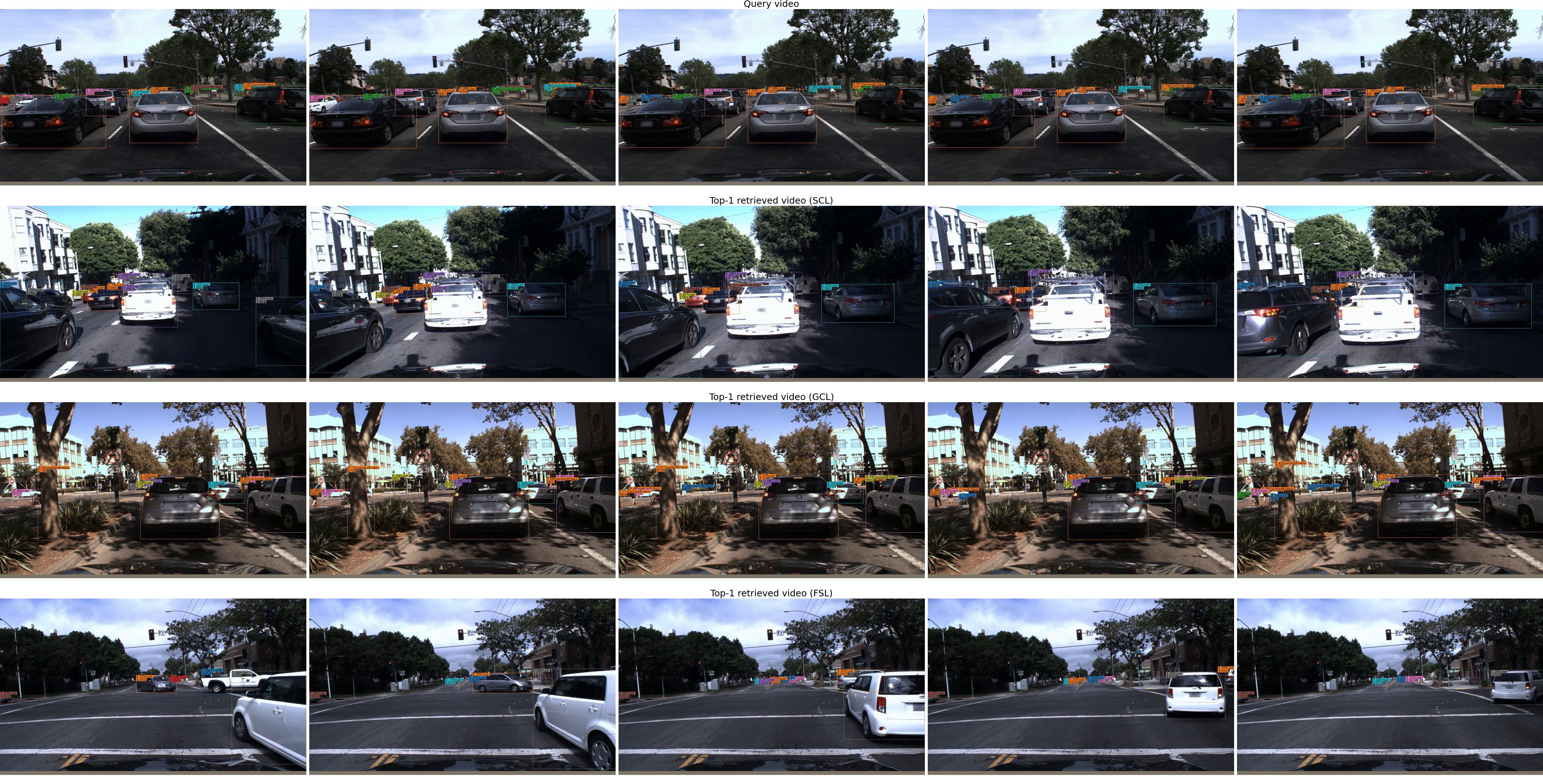}
      \end{minipage}
    \end{tabular}
    \caption{}
  \end{figure*}
  
\begin{figure*}[h]
    \begin{tabular}{c}
      \begin{minipage}[t]{0.95\linewidth}
        \centering
        \includegraphics[width=17.0cm]{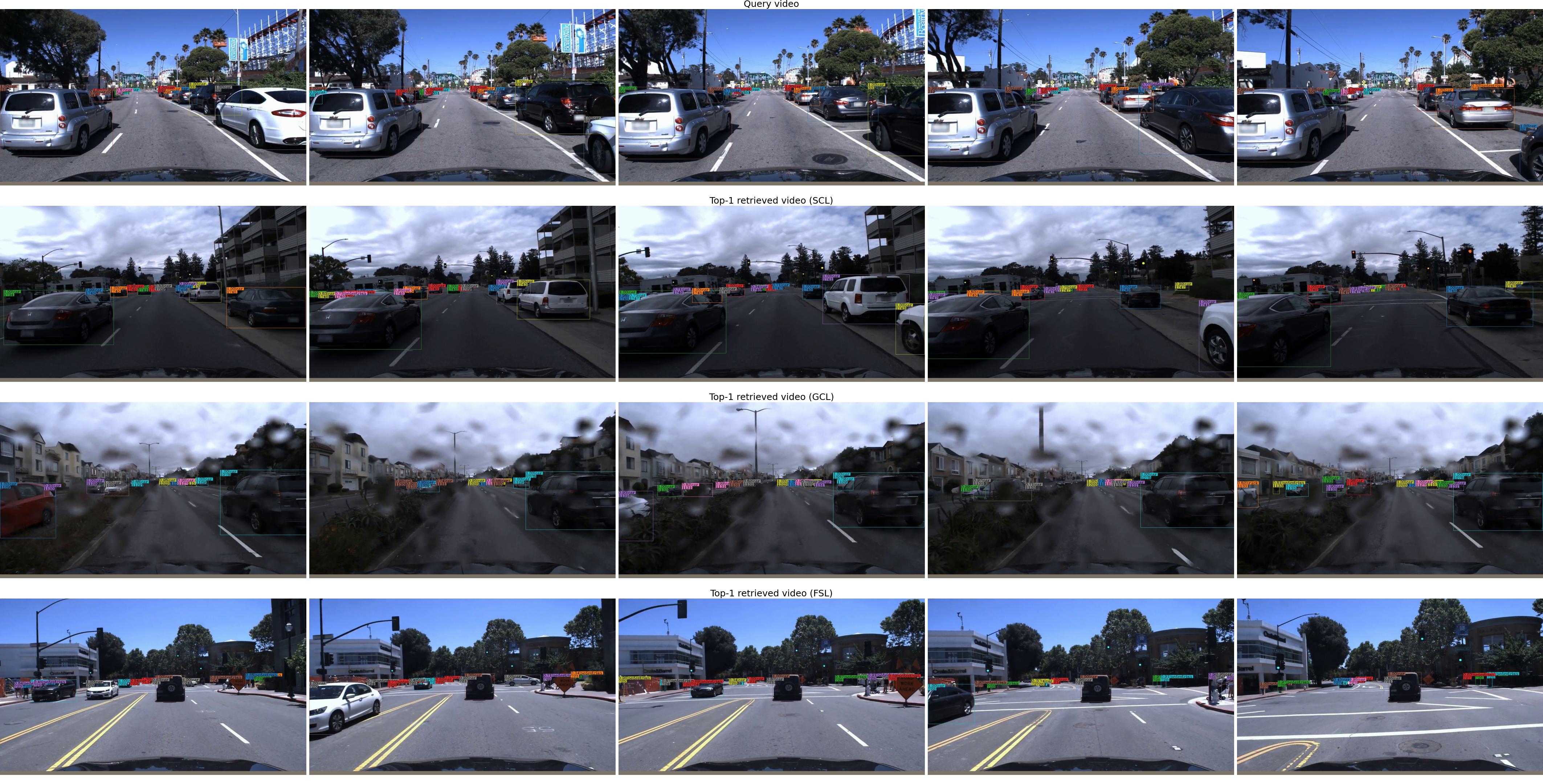}
      \end{minipage}
    \end{tabular}
    \caption{}
  \end{figure*}
  
\begin{figure*}[h]
    \begin{tabular}{c}
      \begin{minipage}[t]{0.95\linewidth}
        \centering
        \includegraphics[width=17.0cm]{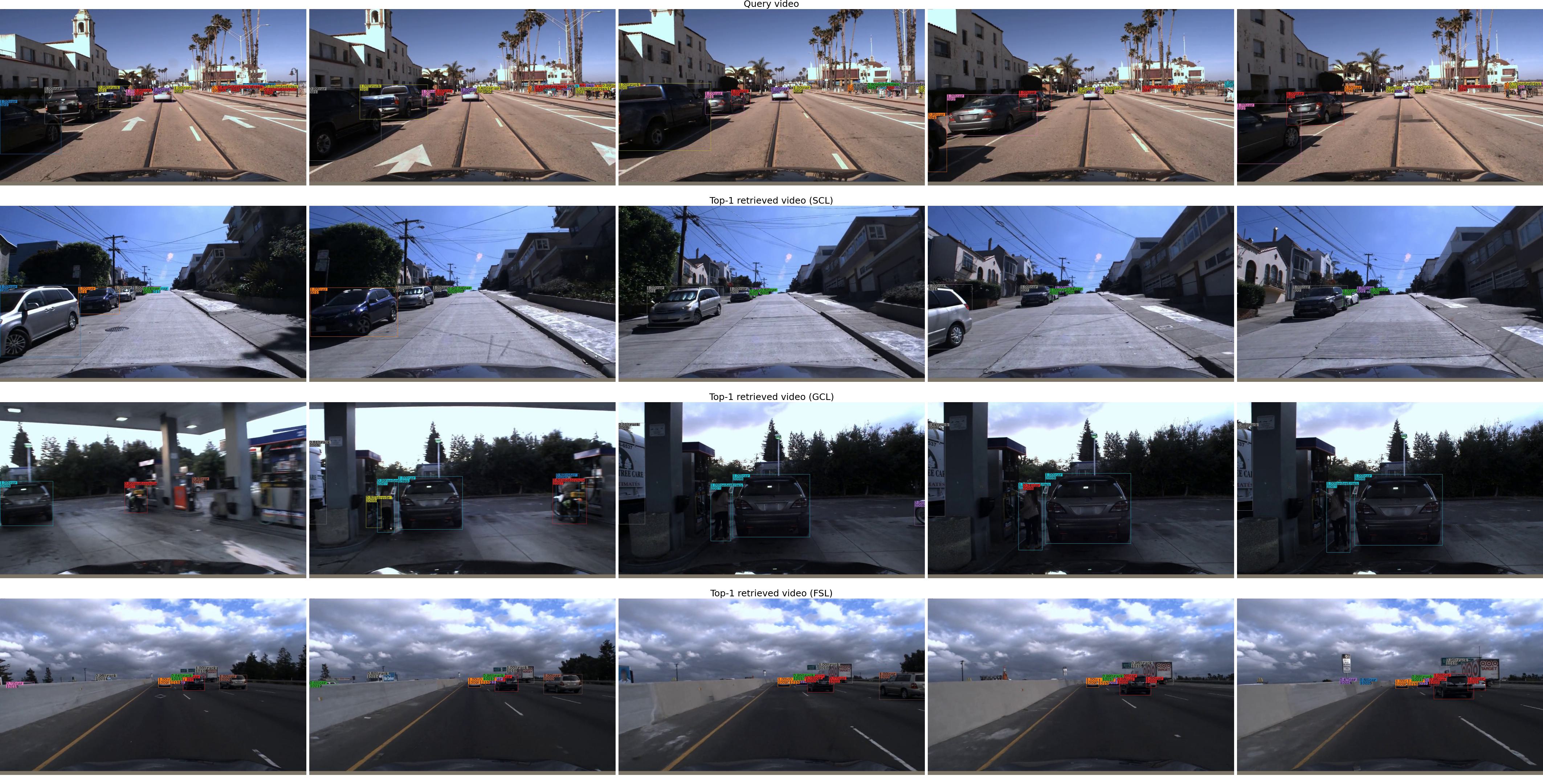}
      \end{minipage}
    \end{tabular}
    \caption{}
  \end{figure*}
  
\begin{figure*}[h]
    \begin{tabular}{c}
      \begin{minipage}[t]{0.95\linewidth}
        \centering
        \includegraphics[width=17.0cm]{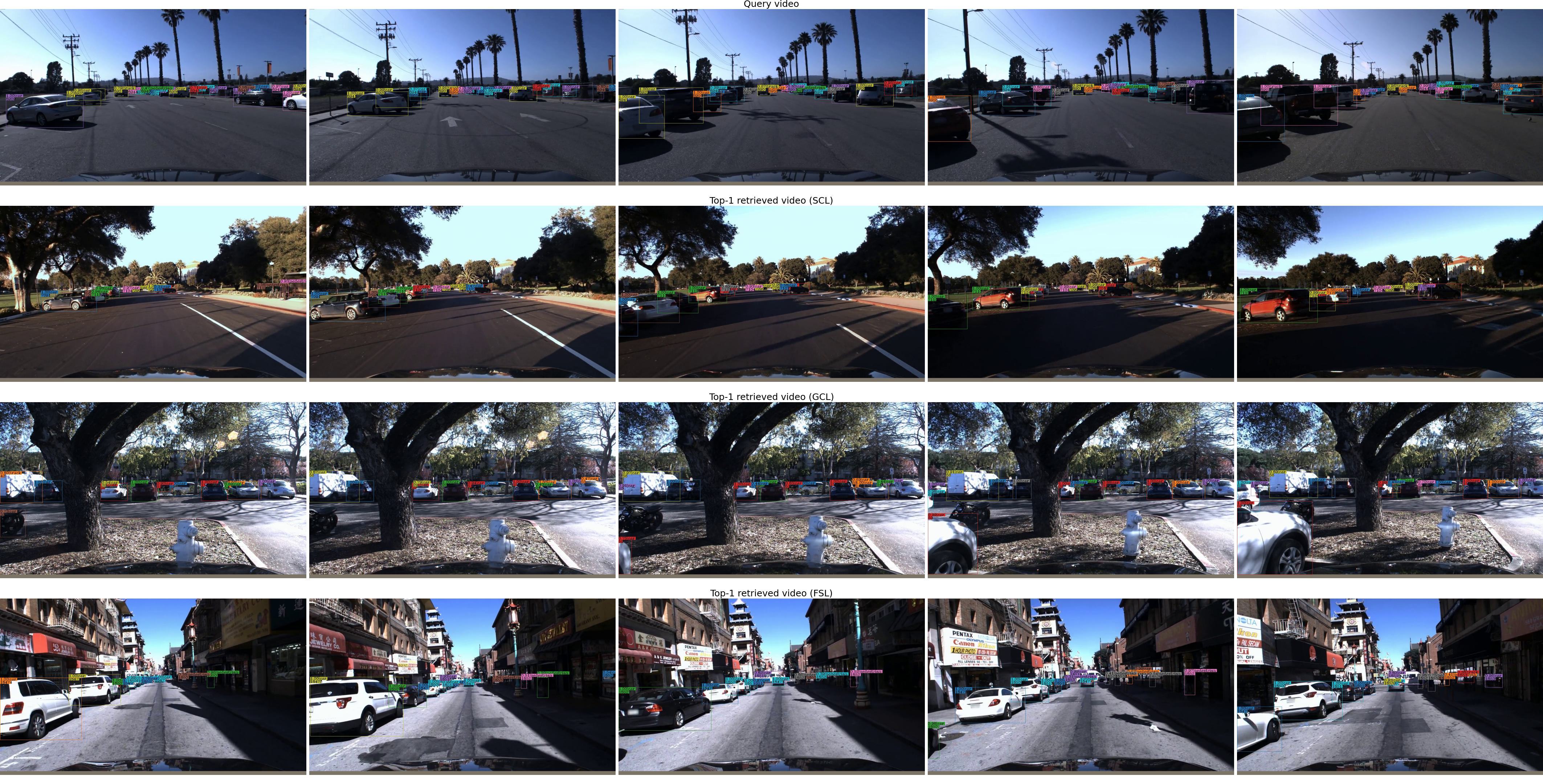}
      \end{minipage}
    \end{tabular}
    \caption{}
  \end{figure*}
  
\begin{figure*}[h]
    \begin{tabular}{c}
      \begin{minipage}[t]{0.95\linewidth}
        \centering
        \includegraphics[width=17.0cm]{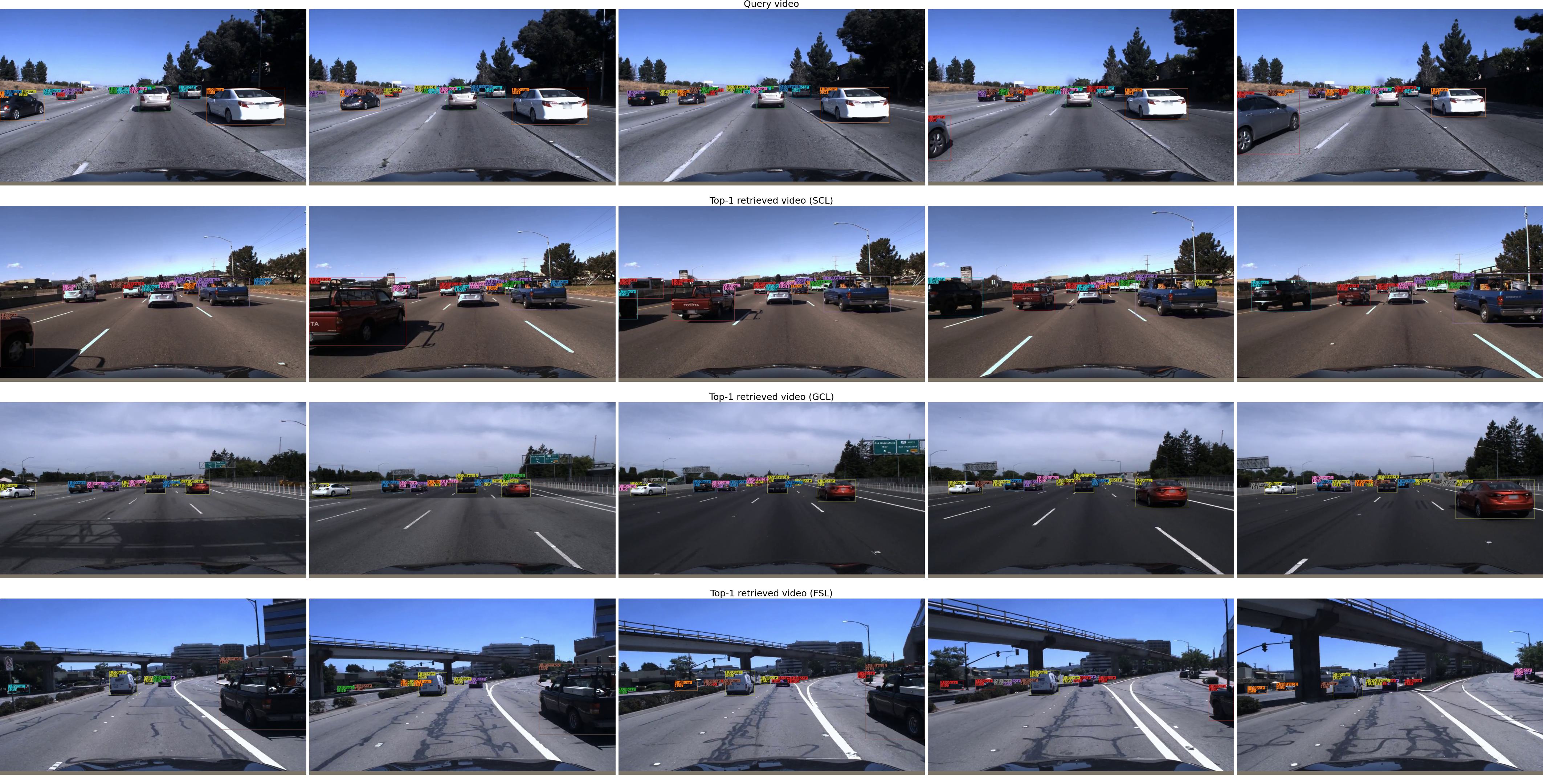}
      \end{minipage}
    \end{tabular}
    \caption{}
  \end{figure*}
  
\begin{figure*}[h]
    \begin{tabular}{c}
      \begin{minipage}[t]{0.95\linewidth}
        \centering
        \includegraphics[width=17.0cm]{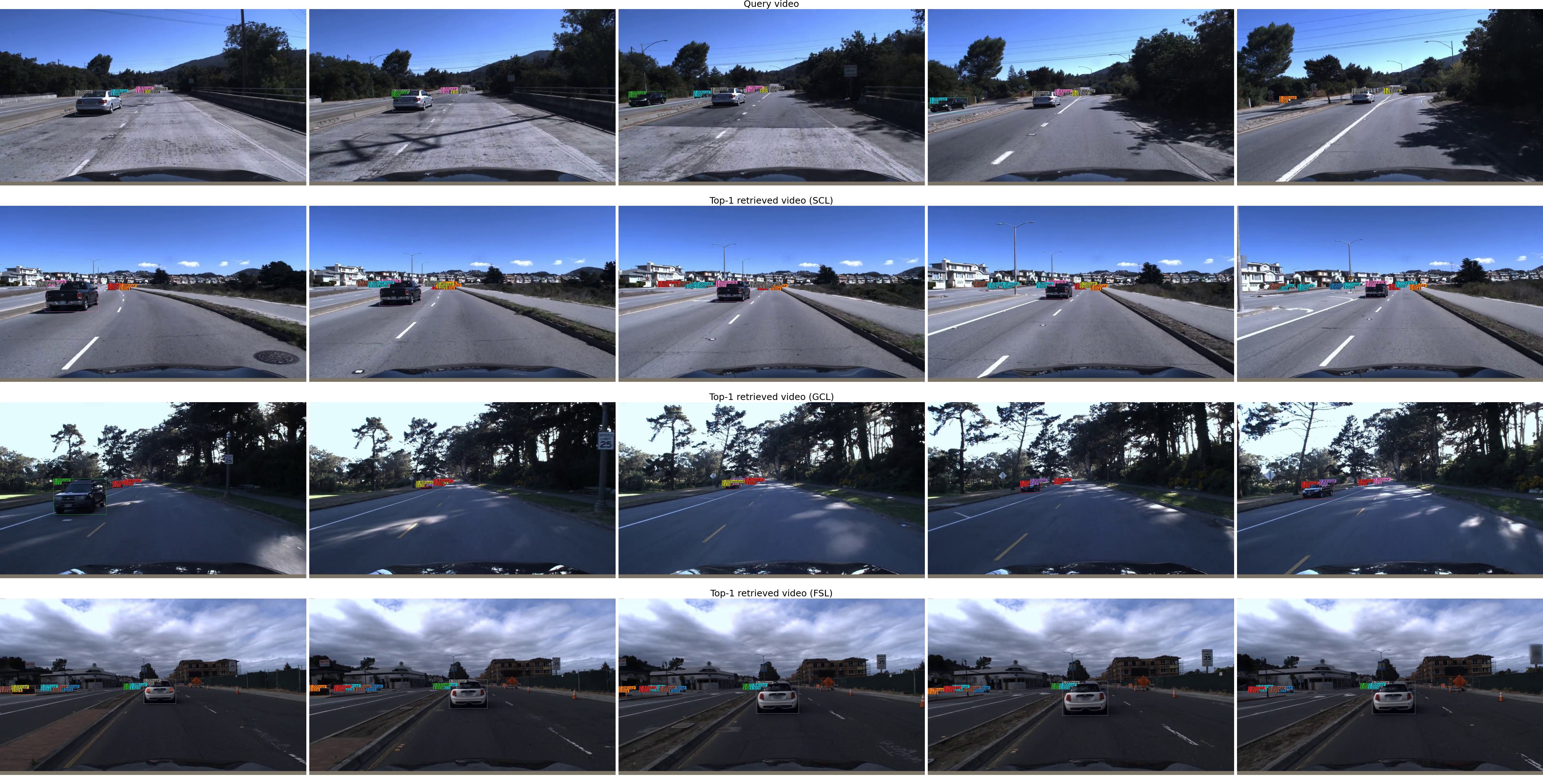}
      \end{minipage}
    \end{tabular}
    \caption{}
  \end{figure*}
  
\begin{figure*}[h]
    \begin{tabular}{c}
      \begin{minipage}[t]{0.95\linewidth}
        \centering
        \includegraphics[width=17.0cm]{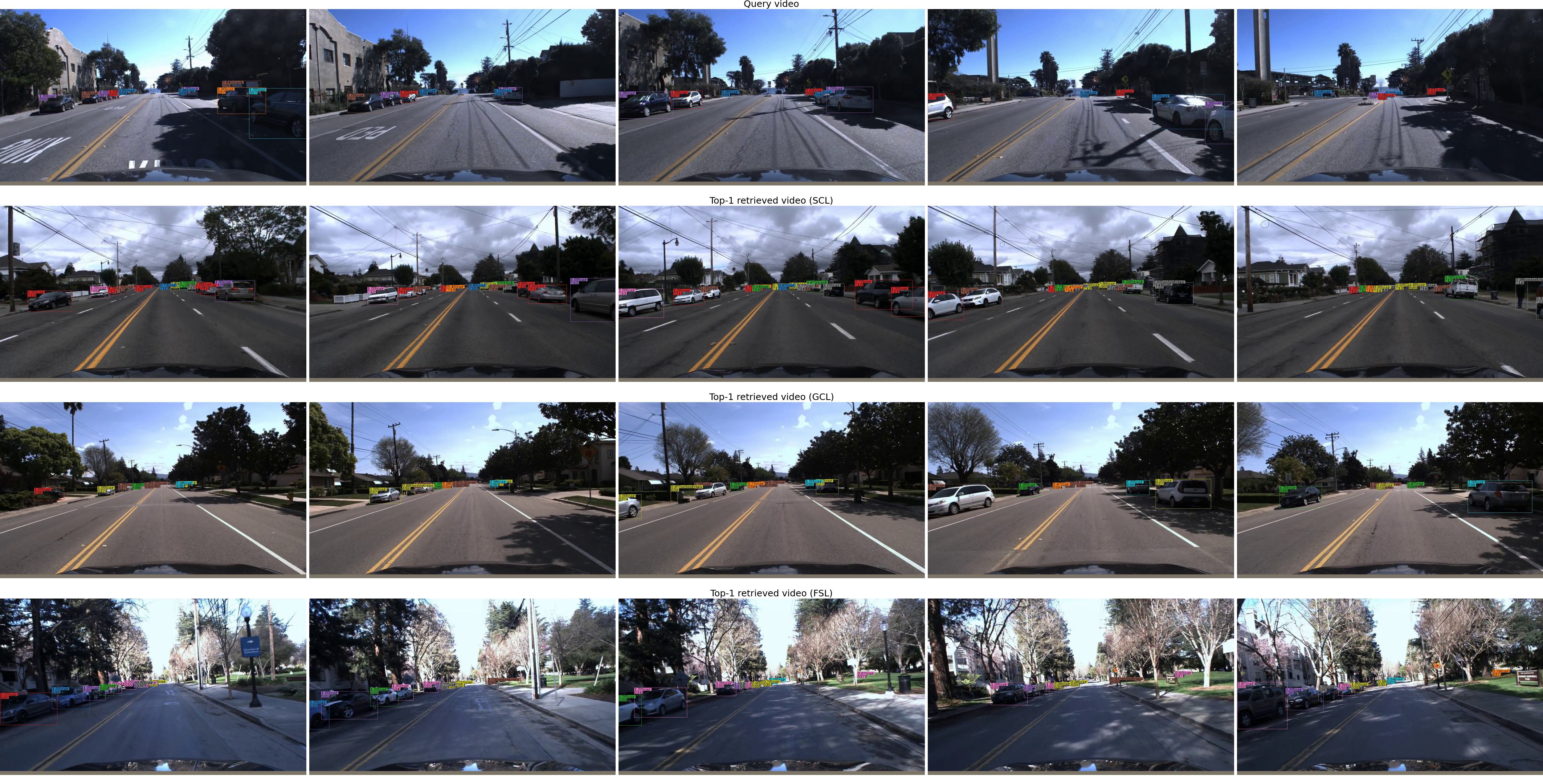}
      \end{minipage}
    \end{tabular}
    \caption{}
  \end{figure*}
  
\end{document}